\documentclass[Afour,sageh,times]{sagej}

\usepackage{moreverb,url}

\usepackage{algorithmicx}
\usepackage{algpseudocode}
\algrenewcommand\algorithmicrequire{\textbf{Input:}}
\algrenewcommand\algorithmicensure{\textbf{Output:}}
\usepackage{xcolor}
\usepackage[linesnumbered,ruled]{algorithm2e}

\newtheorem{theorem}{Definition}[section]
\usepackage{amsmath,amssymb,amsfonts}
\usepackage{bbm}
\usepackage{flushend}
\usepackage[switch]{lineno}

\usepackage[colorlinks, bookmarksopen, citecolor=blue, urlcolor=blue, linkcolor=blue]{hyperref}

\newcommand\BibTeX{{\rmfamily B\kern-.05em \textsc{i\kern-.025em b}\kern-.08em
T\kern-.1667em\lower.7ex\hbox{E}\kern-.125emX}}

\def\volumeyear{2024}
\begin{document}

\runninghead{Wang et al.}

\title{Caging in Time: A Framework for Robust Object Manipulation under Uncertainties and Limited Robot Perception}

\author{Gaotian Wang$^{1*}$, Kejia Ren$^{1*}$, Andrew S. Morgan$^2$ and Kaiyu Hang$^1$}

\affiliation{$^1$Department of Computer Science, Rice University, Houston, TX 77005, USA. 
$^2$The AI Institute, Cambridge, MA 02142, USA.\\
$^*$Equal contribution. \\
Corresponding author: Gaotian Wang, \textit{gw23@rice.edu}
% This work was supported by the National Science Foundation under grant FRR-2240040.
}

% \corrauth{Gaotian Wang,\email{gw23@rice.edu}}

\begin{abstract}
% This paper introduces an innovative approach to robotic manipulation through the concept of temporal caging. Traditional caging methodologies in robotics facilitate manipulation without the necessity for sensing or closed-loop control but often require a special-designed effector or multiple effectors to construct a complete cage around an object, a limitation in scenarios where only a single effector is available. We propose a solution that involves predicting the potential motions of the target object and sequentially building a partial cage that adapts over time. This method effectively achieves the objective of caging the object through a dynamic, time-sensitive process, offering a novel strategy in robotic manipulation for cases with limited effector resources.

Real-world object manipulation has been commonly challenged by physical uncertainties and perception limitations. Being an effective strategy, while caging configuration-based manipulation frameworks have successfully provided robust solutions, they are not broadly applicable due to their strict requirements on the availability of multiple robots, widely distributed contacts, or specific geometries of robots or objects. 

Building upon previous sensorless manipulation ideas and uncertainty handling approaches, this work proposes a novel framework termed \emph{Caging in Time} to allow caging configurations to be formed even with one robot engaged in a task. This concept leverages the insight that while caging requires constraining the object's motion, only part of the cage actively contacts the object at any moment. As such, by strategically switching the end-effector configuration and collapsing it \emph{in time}, we form a cage with its necessary portion active whenever needed. 

We instantiate our approach on challenging quasi-static and dynamic manipulation tasks, showing that \emph{Caging in Time} can be achieved in general cage formulations including geometry-based and energy-based cages. With extensive experiments, we show robust and accurate manipulation, in an open-loop manner, without requiring detailed knowledge of the object geometry or physical properties, or real-time accurate feedback on the manipulation states. In addition to being an effective and robust open-loop manipulation solution, \emph{Caging in Time} can be a supplementary strategy to other manipulation systems affected by uncertain or limited robot perception.

% \rev{In comparison with a closed-loop baseline approach, our experiments show that \emph{Caging in Time} is similarly accurate, while being significantly more robust against various perception uncertainties.}

\end{abstract}

% \keywords{Motion control, Robust manipulation}

\maketitle{}

\section{Introduction}

Object manipulation is a fundamental ability for robots to engage themselves in tasks where physical interactions are expected \citep{Billard19}. While research in this field has seen significant advancements with data-driven frameworks and sensing-enhanced systems \citep{kaelbling2020foundation, yuan2017gelsight, Lee19}, we still do not see many robot systems working robustly around us. Among others, two major reasons are typically observed for this challenge: 1) real-world physical uncertainties are significantly more complex than any lab environment, often rendering certain modeling assumptions invalid \citep{rodriguez2021unstable}, and 2) perception or sensing is almost never good enough for real-world robot deployments. Gaps between modeling assumptions and reality, domain variations, and random occlusions often worsen the perception performance, or even cause a perception system to completely lose track of target objects \citep{bohg17}. As such, approaches fully relying on closed-loop control for robot manipulation have been constantly challenged in real-world applications.

\begin{figure}[!t]
   \centering \includegraphics[width=0.99\columnwidth]{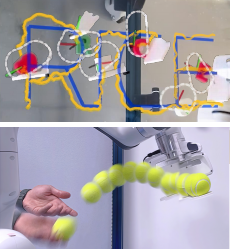}
   \caption{Example object manipulation tasks via \emph{Caging in Time}. \textit{Top}: Object planar pushing to trace ``RICE''. Without sensing feedback, unknown objects were randomly replaced during the manipulation process. The recordings were taken at different times and concatenated to show all objects. \textit{Bottom}: Ball catching on a flat end-effector without any sensing feedback.}

   \label{fig:front}
   \vspace{-10pt}
\end{figure}

% Even with physics-based interactive perception \citep{bohg17}, domain adaptation is still often limited by sensing capabilities.

Unlike approaches that focus on contacts, caging configuration-based manipulation has provided a novel paradigm to significantly reduce perception requirements. More importantly, although it does not aim at accurate control, caging configurations can robustly work by fully ignoring the effect of physical uncertainties, such as seen in grasping, in-hand manipulation, and multi-robot coordination tasks \citep{rodriguez2012caging, song2021herding, bircher_complex_2021}. Concretely, a caging configuration aims at completely constraining all possible configurations of the target object within a known region (cage). The target object is manipulated as the cage moves or deforms, while the configuration of the target object is guaranteed to follow the cage to complete the task \citep{wang2005deformable}. However, it has been a challenge to apply this idea to general manipulation tasks, since forming a caging configuration has very strict requirements on the hardware, such as multi-agent coordination, widely distributed contacts, or specific geometries of the robot or the object to construct such configurations \citep{makita_survey_2017}.

\begin{figure}[ht]
   \centering \includegraphics[width=0.5\textwidth]{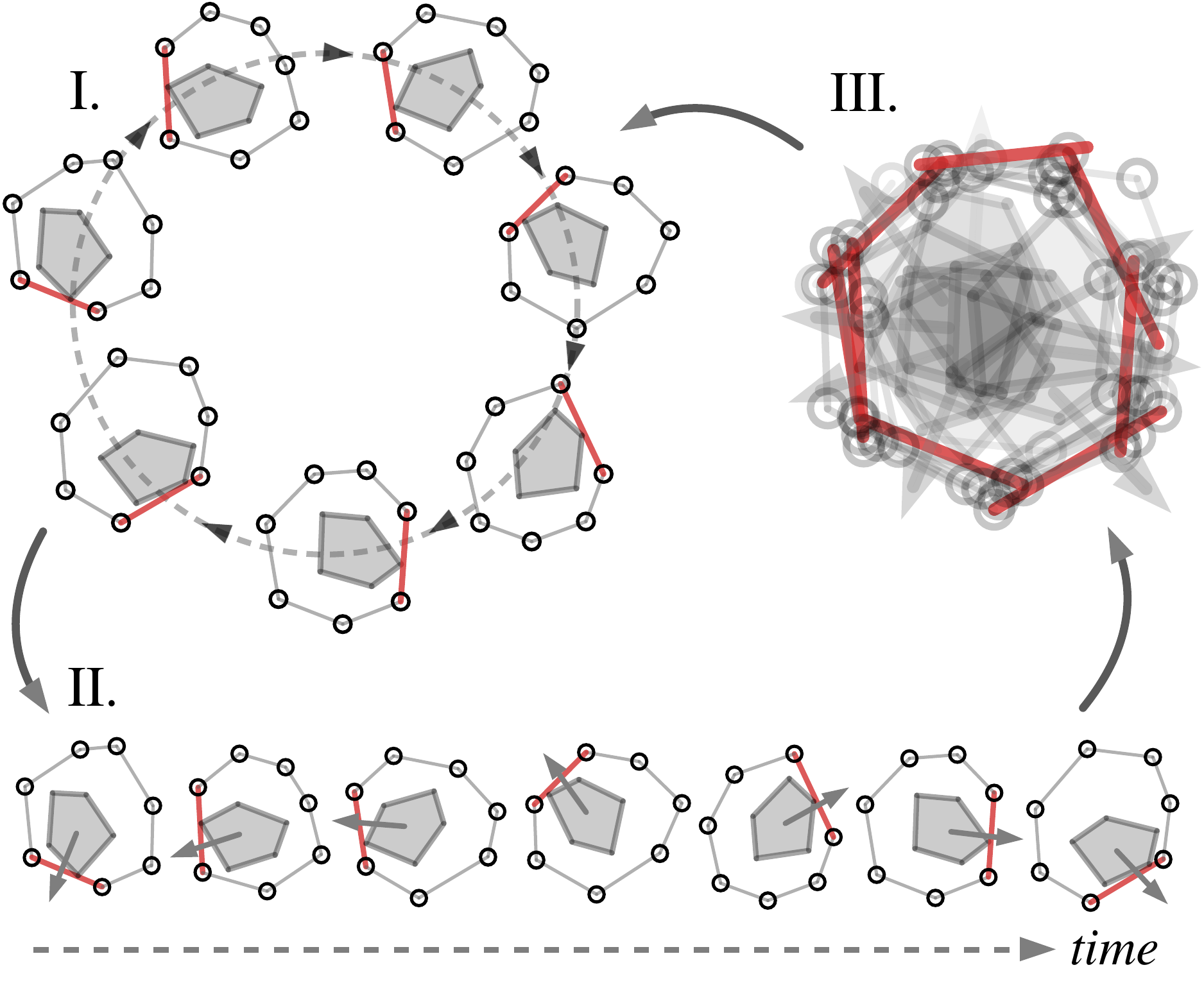}
   \caption{The theory of \emph{Caging in Time} visualized through an example planar pushing task. \textbf{I}: A virtual cage, formed by line-shaped bars (grey), robustly pushes an object through a circular path. \textbf{II}: Along the time dimension, only one bar (red) is effectively making contacts at a time, while other bars seem to be unnecessary. \textbf{III}: If we collapse all configurations of the effective bars (red) through time, a cage is formed, which can be unrolled into time to achieve the task in \textbf{I} with only one bar at a time.}

   \label{fig:cgtp2}
   \vspace{-10pt}
\end{figure}

This work proposes a novel concept, termed \emph{Caging in Time}, to extend caging configuration-based manipulation to more general problems without hardware-specific assumptions. Example applications are shown in Fig.~\ref{fig:front}. The high-level idea of this framework can be explained as follows. In an extreme situation, let us assume we have an object to manipulate and a robot is able to make an infinite number of contacts everywhere on the object. As such, the object is fully caged, and arbitrary manipulation can be achieved by moving all contacts simultaneously. In another extreme situation, let us assume a robot can make one contact with the object at a time, but it is able to switch the contact to other locations infinitely fast so that virtually there are contacts everywhere on the object. Equivalently to the former case, arbitrary manipulation can also be achieved with this virtual cage. Our idea of \emph{Caging in Time} exploits the possibilities between these two extremes: We assume a robot can make one or a few contacts at a time, and it can switch to other contacts fast enough as needed so that, \emph{in time}, it makes a cage. This idea is visualized in Fig.~\ref{fig:cgtp2} via a planar pushing task, where an object is pushed by a virtual cage (grey) with multiple bars through a circular trajectory. Note that, physically and in time, only one bar (red) is effectively needed at a time to complete the task. Furthermore, by collapsing the configurations of the effective bars through time, a cage is formed and can be unrolled in time to complete the task as if a complete cage has always been there.

We instantiated the \emph{Caging in Time} theory on both quasi-static and dynamic manipulation tasks with a real robot. It is worthwhile to note that, the example instantiations showed that \emph{Caging in Time} can be applied on general cage formulations, including geometry-based and energy-based cages. Without any sensing feedback, \emph{Caging in Time} showed guaranteed task success on all experimented tasks, even when the manipulation is physically affected by in-task perturbations and unknown object shape variations. In comparison with a baseline closed-loop control approach, we show that our framework is similarly accurate, while being significantly more robust against perception uncertainties, as enabled by zero reliance on precise sensing feedback.

\emph{Contributions:} The proposed \emph{Caging in Time} concept makes three contributions: 1) providing a planning paradigm for robust robot manipulation that can significantly mitigate the effect of perception uncertainties; 2) broadening the traditional caging configuration-based manipulation to a more general manipulation framework as enabled by strategic sequential robot motions; and 3) offering an option for manipulation without relying on sensing feedback to support robust manipulation in various real-world tasks, especially in scenarios where perception is not reliable.

\emph{Limitations:} The work reported in this paper is the first step towards general applications of the \emph{Caging in Time} concept. With an emphasis on the derivation of the foundations of the theory, this work does not develop algorithms for addressing general manipulation problems via \emph{Caging in Time} skills. Specifically, we identify the following major limitations of the current scope of this work and state them before the details of the proposed theory: 1) as example instantiations of the proposed theory, the reported manipulation algorithms were designed for implementing \emph{Caging in Time} on specific tasks only; 2) manipulation physics or dynamics models are analytically derived for the purpose of explicitly verifying the theory, although simulation or learning-based models can be more efficient; and 3) this work mainly handles perception uncertainties while uncertainties in action execution and certain environmental interactions are not considered.  Nevertheless, we underscore that the proposed \emph{Caging in Time} framework is a complete theory for general manipulation problems as discussed in Sec.~\ref{sec:caging_in_time} and Sec.~\ref{sec:apps}.

\section{Related Works}
\label{sec:related_work}

\emph{Perception Assumptions in Manipulation:} Contact properties, geometric and physical properties of objects, and perfect perception are commonly presumed when modeling manipulation systems \citep{suomalainen2022survey, butepage2019visual}. Alternatively, recent learning-based approaches instead assume training data adequately covers all relevant task variation domains, with perception systems matching those in training environments, e.g., similar camera poses \citep{Lee19, kaelbling2020foundation, andrychowicz2020learning, kroemer2021review}. Meanwhile, interactive perception has significantly reduced the requirements for certain prior knowledge and direct perception \citep{bohg17, hang2021manipulation}, while still assuming reliable perception through some channels for iterative state estimation. Despite demonstrating great performance in the lab, the robustness of many manipulation approaches typically faces significant challenges from unreliable perception in the real world, such as tracking noise, signal latency, and occlusions.

In early works, uncertainty was discussed in object pushing through analytical modeling \citep{lynch1996stable, lynch1995controllability, lynch1999locally, akella_posing_nodate}, enabling robust planar pushing despite uncertainties. While these approaches handled some uncertainties in object-surface interactions, they still required precise object geometry and sensor feedback. Meanwhile, sensorless manipulation \citep{erdmann1988exploration, goldberg1993orienting, akella1997sensorless} provided inspiration for addressing manipulation without any feedback, with applications in orienting planar parts without sensors \citep{akella1998parts, bohringer2000algorithms}. However, these approaches were specifically limited to parts orienting, rather than general manipulation tasks. Later works on motion cones \citep{chavan-dafle2020planar} and in-hand manipulation \citep{bhatt2021surprisingly,holladay2015general} extended these concepts beyond pushing.

Notably, a recent work in robust pushing shares a similar idea with our work through variance-constrained optimization based on belief dynamics, generating stable pushing trajectories also in open-loop \citep{Jankowski25IJRR}. While theoretically rigorous, this approach remains confined to quasi-static pushing scenarios, without addressing its potential applications in broader manipulation contexts.

Therefore, a comprehensive framework that can robustly handle uncertainty even without requiring feedback and apply across diverse manipulation tasks is valuable. Different from most existing works, \emph{Caging in Time} aims to use caging configurations along the time dimension to eliminate the reliance on, hence the negative effect of, unreliable perception. Importantly, our work is not intended to replace existing planning or control methods, but rather to complement them as a supplementary approach to enhance manipulation robustness under unreliable perception.

\emph{Caging Configuration-based Manipulation:} 
% Caging configuration was originated for multi-robot systems in $SE(2)$ to constrain the motion of a target object, so that the motion of the team will determine the motion of the caged object \citep{pereira2004decentralized, sudsang2000new, wang2005deformable}. This concept was later extended to grasping in $2D$, aiming to fix the target object pose within a small region \citep{rodriguez2012caging}. By further modeling the topological relationships between robots and objects, hooking or latching-based manipulation was developed based on caging in topological spaces \citep{stork2013topology}. 
Caging configuration initially emerged as a concept for multi-robot systems in $SE(2)$ to constrain the motion of a target object, enabling robust object transportation \citep{pereira2004decentralized, sudsang2000new, wang2005deformable}, or even herding mobile agents \citep{song2021herding}. This concept was subsequently extended to planar grasping, with the objective of restricting the pose of a target object within a confined region \citep{rimon_caging_1999, rodriguez2012caging, varava2016caging, stork_towards_nodate}, sometimes even with partially observable object geometries \citep{zarubin2013caging}, as well as hooking and latching techniques developed through topological analysis \citep{stork2013topology}. Recently, in-hand manipulation has been enhanced through caging configurations where objects are manipulated via controlled states within deformable cages \citep{komiyama2021position, bircher_complex_2021}. Besides algorithmic advances, the caging concept has also inspired specialized end-effector designs that enhance manipulation through physical embodiment \citep{dong2025cagecoopt, xu2024dynamicsguideddiffusionmodelrobot}. However, the most fundamental limitation of traditional caging is that multiple robots \citep{zhidong_wang_object_2002}, widely distributed contacts \citep{rodriguez2012caging}, or specific geometries of the robots or objects \citep{varava2016caging} are required, making caging a concept not applicable to more general setups.

Beyond complete geometric caging, extended formulations have been proposed including energy-bounded caging to incorporate external forces like gravity  \citep{mahler2016energy, mahler_synthesis_nodate}, and partial caging to relax full enclosure requirements \citep{varava2019partial, varava2016caging} — both demonstrating that complete caging becomes unnecessary with environmental force assistance, broadening the application range for caging. These advancements necessitated practical metrics for escape probability evaluation \citep{varava2019partial, stork_towards_nodate}, with sampling-based algorithms enabling quality assessment through escape path clearance analysis \citep{varava2020free, amato2020caging}. Such quantification methods have evolved beyond traditional caging to evaluate diverse manipulation tasks including pushing, soft object interaction, and multi-object handling \citep{dong2024characterizing, dong_advancing_nodate, dong2023quasi}. 
Despite effectively extending traditional formulations, energy-based partial caging still demands specific robot-object geometry knowledge and maintains complete constraints at each discrete time point, preserving specific end-effector geometry requirements. While robustness prediction metrics have advanced for more general tasks, the fundamental challenges of autonomous caging-based planning and control remain largely unresolved.

Building upon unified geometric and energy-based caging principles, \emph{Caging in Time} departs from the conventional understanding that a cage must be complete at every moment. Instead, we establish a paradigm where a cage can be considered complete when it achieves completeness across the space-time continuum, thus significantly relaxing hardware requirements and expanding the practical applications of caging-based manipulation. Additionally, traditional methods require very complex algorithms to verify caging configurations \citep{varava_free_2021}, which can be even harder for partial cages \citep{makita_evaluation_2015}. As will be seen with \emph{Caging in Time}, since we can, in theory, virtually have our robots or contacts everywhere as needed, caging verification is generally simpler as we only need to make sure that the object does not penetrate the predefined cage boundaries.

% More recently, a specific type of planar dexterous manipulation has been enabled by in-hand caging configurations so that the object is manipulated through controlled energy states inside the cage that deform over time \citep{komiyama2021position,bircher_complex_2021}. With a mobile robot actively trying to escape from a team of other robots, caging configurations have also enabled formation-based motion planning to herd such a robot to a goal region \citep{song2021herding}. Essentially, all existing frameworks aim at first constructing a cage to enclose the target object, and then maintain the cage while moving or deforming it to achieve a manipulation task. 

\emph{Theories for Uncertainty Handling}: Our approach draws inspiration from foundational theoretical frameworks that address uncertainties through diverse mathematical formulations across control theory, motion planning, and decision-making. Belief space planning transforms state estimation into probability distributions for decision-making under uncertainty \citep{jr_belief_nodate, kurniawati_sarsop_nodate}. POMDPs formalize partial observability by optimizing over belief states \citep{kaelbling_planning_1998, silver_monte-carlo_nodate}. Contraction theory provides stability guarantees through analysis of convergence properties \citep{lohmiller_contraction_1998, manchester_control_2017}, while reachability analysis enables formal safety verification by computing attainable states \citep{mitchell_time-dependent_2005, altho_reachability_nodate}. LQR trees combine optimal control with sampling-based planning for stabilizing controllers \citep{tedrake_lqr-trees_2010, majumdar_funnel_2017}, and set-based control ensures invariance properties for worst-case scenarios \citep{rakovic_invariant_2005}.

While all these approaches can effectively address uncertainties in different ways, they still rely on the assumption that perception feedback or certain prior geometric knowledge of the tasks is always, or at least partly, available. Inspired by these prior works and with an aim to address their limitations in contact-rich manipulation tasks, our proposed \emph{Caging in Time} synthesizes their insights, including those in state representations and uncertainty-aware planning, from these fundamental theories into a framework that focuses on large perception uncertainties, in order to enable robust manipulation skills with practical real-world implementations.
\section{Preliminaries}
\label{sec:preliminaries}

In this section, we first introduce notations in Sec.~\ref{sec:rep_and_not} for traditional caging definitions, and then in Sec.~\ref{sec:general} extend the notations to more general task spaces to enable the derivation of our proposed \textit{Caging in Time} framework.

\subsection{Traditional Caging as Object Closure}
\label{sec:rep_and_not}
\label{sec:obj_closure}

\begin{figure}[t]
   \centering
   \includegraphics[width=0.45\textwidth]{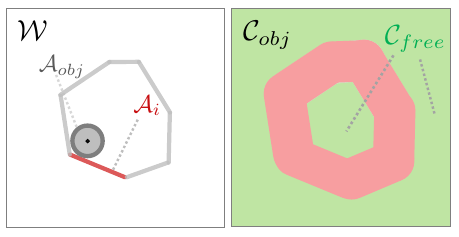}

   \caption{An illustration of the workspace $\mathcal{W}$ (left)  and the C-space $\mathcal{C}_{obj}$ (right) of a circular object surrounded by seven line robots. \textit{Left}: The red line is the geometry of the $i$-th robot $\mathcal{A}_i$ and the solid grey circle is the geometry of the object $\mathcal{A}_{obj}$. \textit{Right}: The red region is all the object's configurations at which the object collides with a robot and the green region is the object's free C-space $\mathcal{C}_{free} \subset \mathcal{C}_{obj}$. }
   \label{fig:cg1}
   \vspace{-10pt}
\end{figure}

In the scenario of an object caged by multiple robots, we denote the configuration space (C-space) of an object by $\mathcal{C}_{obj}$ and the configuration of an object (e.g., position and orientation) by $c_{obj} \in \mathcal{C}_{obj}$.
The C-space and the configuration of the $i$-th robot are denoted by $\mathcal{C}_{i}$ and $c_i \in \mathcal{C}_i$, respectively.
The object and all the robots share a common workspace $\mathcal{W}$, where $\mathcal{A}_{obj}, \mathcal{A}_i \subset \mathcal{W}$ are the geometries of the object and of the $i$-th robot in the workspace.
As shown in Fig.~\ref{fig:cg1}, the free C-space of the object $\mathcal{C}_{free} \subset \mathcal{C}_{obj}$ can be defined as the set of all possible states where the object does not collide with any robot:
\begin{equation}
    \mathcal{C}_{free} := \left\{c_{obj} \in \mathcal{C}_{obj} \mid \forall i : \mathcal{A}_{obj} (c_{obj}) \cap \mathcal{A}_i(c_i) = \emptyset \right\}
\end{equation}

The traditional definition of caging, initially termed object closure, was first introduced by \citep{zhidong_wang_object_2002} as a condition for an object to be trapped by robots.
In other words, the caging condition is met when there is no viable path for the object to move from its current configuration to a configuration infinitely far away.
\begin{figure}[t]
   \centering
   \includegraphics[width=0.45\textwidth]{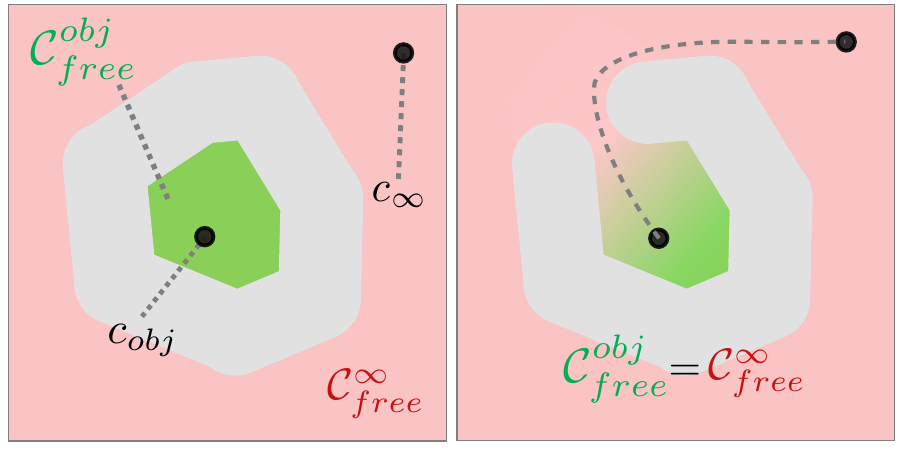}
   \caption{An illustration of the traditional caging condition. \textit{Left}: The green region and the red region represent $\mathcal{C}_{free}^{obj}$ and $\mathcal{C}_{free}^\infty$, respectively. The object is caged in this case since $\mathcal{C}_{free}^{obj}$ and $\mathcal{C}_{free}^\infty$ are not connected. \textit{Right}: A non-caging scenario where $\mathcal{C}_{free}^{obj}$ and $\mathcal{C}_{free}^\infty$ become connected. The object may escape following the trajectory shown by the dashed line.}
   \label{fig:cg2}
   \vspace{-5pt}
\end{figure}
As depicted in Fig.~\ref{fig:cg2}, given a point infinitely far away from the current $c_{obj}$ in the  C-space $c_\infty \in \mathcal{C}_{obj}$, two distinct sets $\mathcal{C}_{free}^{obj}, \mathcal{C}_{free}^\infty \subset \mathcal{C}_{free}$ are defined as follows. 
$\mathcal{C}_{free}^{obj}$ is the largest subset of $\mathcal{C}_{free}$ within which every configuration $c$ is reachable from the current configuration $c_{obj}$ through a collision-free path; similarly, $\mathcal{C}_{free}^{\infty}$ is the largest subset of $\mathcal{C}_{free}$ within which every $c$ can reach $c_\infty$:
\begin{equation}
    \begin{aligned}
        \mathcal{C}_{free}^{obj} &=\left\{c \in \mathcal{C}_{free} \mid \operatorname{connected}\left(c, c_{obj}\right)\right\},\\
        \mathcal{C}_{free}^\infty &=\left\{c \in \mathcal{C}_{free} \mid \operatorname{connected}\left(c, c_{\infty}\right)\right\}
    \end{aligned}
\end{equation}

The caging condition is met if and only if:
\begin{equation}
\label{eq:caging}
\left\{\begin{array}{l}
\mathcal{C}_{free}^{obj} \neq \emptyset \\
% \mathcal{C}_{ {free}^{obj}} \cap \mathcal{C}_{ free}^{\infty} = \emptyset
\mathcal{C}_{free}^{obj}  \cap \mathcal{C}_{free}^\infty = \emptyset
\end{array}\right.
\end{equation}
where the first criterion ensures the existence of a caging configuration, and the second condition guarantees no feasible path connects $\mathcal{C}_{free}^{obj}$ and $\mathcal{C}_{free}^\infty$. This traditional concept of caging serves as the foundation for our proposed \emph{Caging in Time} framework.

\subsection{Generalized Representations}
\label{sec:general}

To enable the derivation of the \emph{Caging in Time} theory, we now generalize the representations and notations from configuration spaces to state spaces. This allows for more comprehensive descriptions of manipulation tasks and accommodates various dynamic scenarios and uncertainties, as opposed to the traditional quasi-static configuration-based caging definitions.

We denote the state space of an object by $\mathcal{S}_{obj}$, which encompasses a wider range of properties than $\mathcal{C}_{obj}$. The state of an object at time $t$ is represented by $\mathbf{q}_t \in \mathcal{S}_{obj}$. Depending on the specific manipulation task, $\mathbf{q}_t$ may include not only the configuration (e.g., position and orientation) but also velocity, acceleration, or other relevant properties.

Furthermore, often in real-world systems, the state of the object at time $t$ is not exactly known due to perception uncertainties or modeling simplifications.
% , or an open-loop state estimator
 To account for such uncertainties, we define the \textbf{Potential State Set} (PSS) of an object to be the set consisting of all possible states of the object at time $t$, denoted by $\mathcal{Q}_t \subset \mathcal{S}_{obj}$.

\section{Caging in Time}
\label{sec:caging_in_time}

In this section, we will formally introduce the definition of our proposed \emph{Caging in Time} theory. With as few as only a single robot interacting with the object, \emph{Caging in Time} verifies that the object's state remains caged while being manipulated, thereby enabling open-loop manipulation and guaranteeing the desired movement of the object.
% \emph{Caging in Time} only requires a single robot to interact with the object being manipulated.
% \rev{It} only requires a single robot to interact with the object being manipulated.
As an essential component of our theory, we need to predict the bounded motions of the object, which involves propagating the PSS over time, as will be introduced in Sec.~\ref{sec:prop_pcs}.
Then, we will formally define \emph{Caging in Time} in Sec.~\ref{sec:sub_caging_in_time}. 

For more intuitive illustrations of the proposed concepts, figures in this section are sketched in 2D spaces. However, it is important to emphasize that $\mathbf{q}_t$ can take any form and dimension as required by the specific manipulation task. The 2D visualizations are chosen for visual simplicity and do not limit the generality of the state space concept.

\subsection{Propagation of PSS}
\label{sec:prop_pcs}
At time $t$, for each specific state of the object in the PSS $\mathbf{q}_t \in \mathcal{Q}_t$, we denote the set of all allowed motions that transit the object's state by $\mathcal{V}_{\mathbf{q}_t}$.
We define the notation $\mathrm{T}_x \mathcal{M}$ to represent the tangent space of a manifold $\mathcal{M}$ at a point $x\in \mathcal{M}$ in this manifold.
Note that $\mathcal{V}_{\mathbf{q}_t} \subset \mathrm{T}_{\mathbf{q}_t} \mathcal{S}_{obj}$, i.e., $\mathcal{V}_{\mathbf{q}_t}$ lies in the tangent space of the object's state space at the state $\mathbf{q}_t$.
Based on the tangent bundle theory \citep{jost_riemannian_2011}, we define the motion bundle at time $t$ in Eq.~\eqref{eq: QV_t}, 
which is the set consisting of all possible state-motion pairs of the object:
\begin{equation}
\label{eq: QV_t}
    \mathcal{QV}_t := \{(\mathbf{q}_t, \mathbf{v}_t) \mid \mathbf{q}_t \in \mathcal{Q}_t, \mathbf{v}_t \in \mathcal{V}_{\mathbf{q}_t}\}
\end{equation}

We need to define a propagation function $\pi: \mathcal{QV}_t \mapsto \mathcal{S}_{obj}$ to predict the object's state $\mathbf{q}_{t+1}$ at the next time step given a possible state $\mathbf{q}_t$ and a motion $\mathbf{v}_t$, i.e., $\mathbf{q}_{t+1} = \pi(\mathbf{q}_t, \mathbf{v}_t)$.

\begin{figure}[t]
   \centering
   \includegraphics[width=0.45\textwidth]{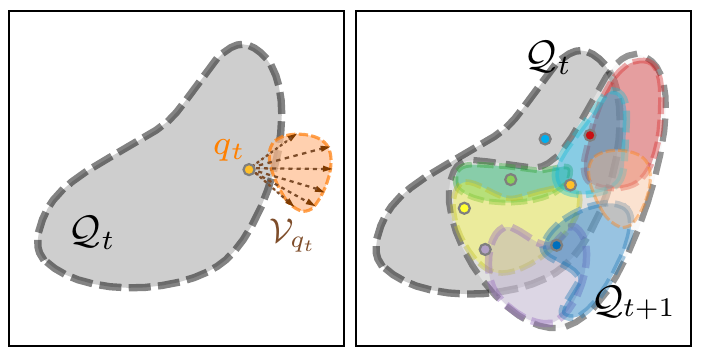}
   \caption{An illustration of the PSS propagation. \textit{Left}: The propagation for a single state  $\mathbf{q}_t \in \mathcal{Q}_t \subset \mathcal{S}_{obj}$. With a set of possible motions $\mathcal{V}_{\mathbf{q}_t}$, $\mathbf{q}_t$ can be propagated to a set of different states (the orange region). \textit{Right}: The propagation from the entire PSS $\mathcal{Q}_{t}$ to $\mathcal{Q}_{t+1}$ by propagating all points in $\mathcal{Q}_{t}$. The different colors of regions in $\mathcal{Q}_{t+1}$ illustrate the propagation from multiple different $\mathbf{q}_t \in \mathcal{Q}_t$.}
   \label{fig:cgt1}
   \vspace{-10pt}
\end{figure}

However, the above propagation function only predicts a single state for the next time step.
In our \emph{Caging in Time} framework, we need to know all the possible states propagated from the previous step, i.e., the PSS $\mathcal{Q}_{t+1}$.
To this end, we extend the propagation function $\pi$ to another function $\Pi: 2^{\mathcal{QV}_t} \mapsto 2^{\mathcal{S}_{obj}}$ whose domain is the power set of the motion bundle $\mathcal{QV}_t$.
As illustrated in Fig.~\ref{fig:cgt1}, it is used to obtain the entire set of all possible states at time $t+1$ by propagating every possible $\mathbf{q}_t$:
\begin{equation}
    \label{piact}
    \begin{aligned}
        \mathcal{Q}_{t+1} &= \Pi(\mathcal{QV}_t) \\
        &= \{\mathbf{q}_{t+1} = \pi(\mathbf{q}_t, \mathbf{v}_t) \mid (\mathbf{q}_t, \mathbf{v}_t) \in \mathcal{QV}_t \}
    \end{aligned}
\end{equation}
\begin{figure}[t]
   \centering
   \includegraphics[width=0.45\textwidth]{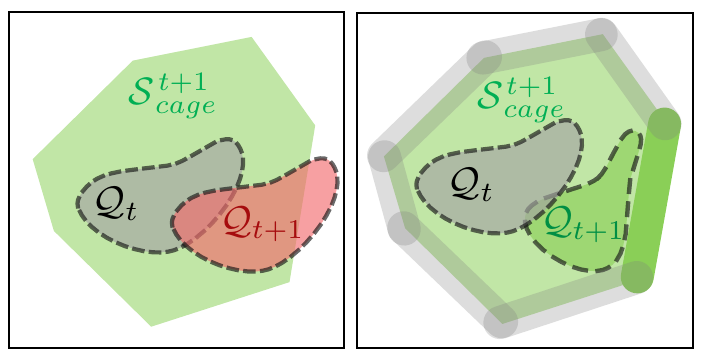}
   \caption{A 2D pushing example showing how the PSS propagation is influenced by a robot action.
    \textit{Left}: A scenario where $\mathcal{Q}_t$ propagates to $\mathcal{Q}_{t+1}$ without robot intervention. $\mathcal{Q}_{t+1}$ will exceed the cage region $\mathcal{S}_{cage}^{t+1}$. \textit{Right}: With a well-selected robot push starting from the pose shown by the green bar towards the center of $\mathcal{S}_{cage}^{t+1}$, $\mathcal{Q}_{t+1}$ will deform to a different shape and stay inside $\mathcal{S}_{cage}^{t+1}$. In this example, the workspace and the object's state space (i.e., 2D position of the object) share the same space $\mathbb{R}^2$, so we overlay the robot pushes (grey bars) onto the object's state space for visualization. }
   \label{fig:cgt2}
   \vspace{-10pt}
\end{figure}

\subsection{Caging in Time}
\label{sec:sub_caging_in_time}

Unlike the traditional caging introduced in Sec.~\ref{sec:obj_closure}, our \emph{Caging in Time} framework only requires one robot to manipulate the object.
The cage is formed over time by switching the single robot to a different configuration and interacting with the object differently at each time step.
Fig.~\ref{fig:cgt2} shows an example where a robot pusher can push an object from various possible initial locations relative to the object (marked by the grey bars on the right).
At each time step, by predicting the object's possible motion via the propagation function $\Pi$ defined in Sec.~\ref{sec:prop_pcs}, the robot needs to figure out which candidate push to select (e.g., the green bar in the right figure of Fig.~\ref{fig:cgt2}), to prevent the object from escaping.
As such, the object can always be confined to a bounded region as the robot pusher switches to different positions over time and pushes the object in different directions.

Specifically, for every time step $t$, we want to confine the object's state to be always inside a region $\mathcal{S}_{cage}^t \subset \mathcal{S}_{obj}$ in its state space, and we term this region the ``cage region''.
The ``cage region'' $\mathcal{S}_{cage}^t$ does not have to be stationary; it can deform and displace depending on the manipulation task it entails.
Moreover, a set of candidate robot actions $\mathcal{U}$ needs to be predefined based on the context of the task.
For example, in the planar pushing manipulation scenario shown in Fig.~\ref{fig:cgt2}, $\mathcal{U}$ can be a set of robot-pushing actions defined by different starting positions and pushing directions of the robot pusher.
Such robot actions will affect the possible motion of the object.
For example, while the robot is interacting with the object through contact, the object will not be able to move in the directions that penetrate the robot.
In other words, for a specific object state $\mathbf{q}_t \in \mathcal{S}_{obj}$, the allowed object motions $\mathcal{V}_{\mathbf{q}_t}$ will be influenced by the robot action $u \in \mathcal{U}$.
We use a function $U$ to represent such influence by 
\begin{equation}
\label{eq:U}
    \mathcal{V}_{\mathbf{q}_t} = U(\mathbf{q}_t, u) \subset \mathrm{T}_{\mathbf{q}_t} \mathcal{S}_{obj}
\end{equation}

Applying a robot action $u$ may cause the object to move in a direction towards the outside of the cage region $\mathcal{S}_{cage}^{t+1}$ at the next time step.
To guarantee that the object's state is always inside $\mathcal{S}_{cage}^t$, for each time step $t = 1, \cdots, T$, there must exist some robot action $u_t \in \mathcal{U}$ that causes all possible states of the object to be contained in $\mathcal{S}_{cage}^{t+1}$ after the robot action is executed.
As such, the definition of \emph{Caging in Time} is formally given as follows.

\begin{theorem}
For a time-varying cage $\mathcal{S}_{cage}^t$ in an object's state space where $t = 1, \cdots, T$, \emph{Caging in Time} is achieved if and only if: $\forall t = 0, 1, \cdots, T-1$, $\exists u_t \in \mathcal{U}$ such that
\begin{equation}
    \begin{aligned}
        \mathcal{QV}_t &= \{(\mathbf{q}_t, \mathbf{v}_t) \mid \mathbf{q}_t \in \mathcal{Q}_t, \mathbf{v}_t \in U(\mathbf{q}_t, u_t)\}, \\
        \mathcal{Q}_{t+1} & = \Pi(\mathcal{QV}_t) \subset \mathcal{S}_{cage}^{t+1}\\
    \end{aligned}
\end{equation}
\label{def:Caging_in_Time}
\end{theorem}
It is worth noting that \emph{Caging in Time} does not guarantee the existence of such robot actions $\{u_t\}_{t=0}^{T-1}$ to cage the object.
However, if a sequence of robot actions $\{u_t\}_{t=0}^{T-1}$ is verified to meet the \emph{Caging in Time} condition, it is guaranteed that the object's state always lies inside the time-varying cage $\mathcal{S}_{cage}^t$.

The process of determining whether \emph{Caging in Time} is achieved is detailed in Alg.~\ref{alg:caging_in_time}.
When the \emph{Caging in Time} condition is met, the object is guaranteed to be caged inside a cage region $\mathcal{S}_{cage}^t$ through a certain sequence of robot actions in an open-loop manner.
If we apply this action sequence, the object will never escape even without any sensing feedback.
As aforementioned, the cage region $\mathcal{S}_{cage}^t$ does not have to be static; it is allowed to change over time.
For example, if $\mathcal{S}_{cage}^t$ is moved to follow a certain trajectory in the state space of the object over time, the object being caged in time will be able to follow the same trajectory of $\mathcal{S}_{cage}^t$.

\emph{Verification of Action Feasibility:}
In practice, the robot action $u_t$ may not be realizable due to the speed limit or reachability of the robot, which varies on different robot platforms.
For example, between the executions of consecutive actions $u_{t-1}$ and $u_t$, the object may keep moving if without certain assumptions, such as quasi-static motions.
This requires the robot to switch to the next action execution fast enough, which may exceed the hardware's speed limit. 
To this end, at Line $3$ of Alg. \ref{alg:caging_in_time}, to achieve \emph{Caging in Time} under a real-world setting, it is necessary to verify the feasibility of the action $u_t$ by taking into account the capability of the actual hardware.

\begin{algorithm}[t]
\SetKwInput{KwData}{Input}
\SetKwInput{KwResult}{Output}

\caption{The Verification of \emph{Caging in Time}}
\label{alg:caging_in_time}

\KwData{The time-varying cage region $\mathcal{S}_{cage}^t \subset \mathcal{S}_{obj}$}, a sequence of robot actions $\{u_t \in \mathcal{U}\}$
\KwResult{The object is caged in time or not}

$\mathcal{Q}_0 \gets $ initial PSS of the object\\
  \For{$t = 0, 1, \cdots, T-1$}{
    \If{ not $\Call{Feasible}{u_t}$}{\Return{false}}
    $\mathcal{QV}_t \gets \{(\mathbf{q}_t, \mathbf{v}_t) \mid \mathbf{q}_t \in \mathcal{Q}_t, \mathbf{v}_t \in U(\mathbf{q}_t, u_t)$\}\\
    \Comment{Construct the Motion Bundle}\\
    $\mathcal{Q}_{t+1} \gets \Pi(\mathcal{QV}_t)$ \hfill 
    \Comment{Propagation of PSS via $\Pi$}

    \If{$\mathcal{Q}_{t+1} \not\subset \mathcal{S}_{cage}^{t+1}$}{
        \Return{false}
    }
  }
  \Return{true}
\end{algorithm}
\section{Quasi-static Tasks}
\label{sec:planar_pushing}

In this section, we instantiate the \emph{Caging in Time} theory on a quasi-static planar pushing problem.
To facilitate the instantiation, we also develop relevant tools for propagating the PSS of the object in a 2D space and generating open-loop robot pushing actions to cage the object. 
With the \emph{Caging in Time} framework applied to the planar pushing problem, we can enable a single robot to push an object of unknown shape to follow certain trajectories in an open-loop manner, without requiring sensory feedback, exact geometric information, or any other physical properties of the object. This is an instantiation of \emph{Caging in Time} with a geometry-based cage.

\subsection{Problem Statement}
\label{sec:prob_stat}

The robot is tasked to push an object on a 2D plane.
To generalize over different shapes of the object without requiring exact geometric modeling, we represent the object's geometry by a simplified bounding circle with radius $r$, i.e., $\mathcal{A}_{obj}(\mathbf{q}) = \left\{\mathbf{p} \in \mathbb{R}^2 \mid \lVert \mathbf{p} - \mathbf{q} \rVert \leq r \right\}$ where $\mathbf{q}$ denotes the position of the object.
The radius $r$ is chosen based on the size of the object such that the bounding circle covers the actual shape of the object. The robot is equipped with a fixed-length line pusher, represented by a line segment $\mathcal{L}$ in the 2D space.
The object is considered to be in collision with the line pusher when $\mathcal{L}$ intersects the bounding circle of the object, i.e., the distance from $\mathbf{q}$ to the line segment $\mathcal{L}$ is not greater than the radius $r$.

While the configuration space of a quasi-static 2D object is $SE(2)$, requiring both the position and orientation of the object, our use of a bounding circle allows us to simplify the object's state space to its position only: $\mathcal{S}_{obj} \subset \mathbb{R}^2$.
As such, the state of the object becomes $\mathbf{q}_t = (x_t, y_t) \in \mathcal{S}_{obj}$. Therefore, both the PSS $\mathcal{Q}_t$ and the cage $\mathcal{S}_{cage}^t$ can be conceptualized as regions within this 2D plane.
In this specific planar pushing task, we further define the \textbf{Potential Object Area} (POA), $\mathcal{P}_t \subset \mathbb{R}^2$, to be the set consisting of all the points in the workspace that are possibly occupied by the object's geometric shape:
\begin{equation}
    \mathcal{P}_t := \bigcup_{\mathbf{q}_t \in \mathcal{Q}_t} \mathcal{A}_{obj}(\mathbf{q}_t)
\end{equation}

The ``cage region'' $\mathcal{S}_{cage}^t \subset \mathcal{S}_{obj}$ is then given in order to cage the object's geometric coverage $\mathcal{P}_t$.
Specifically, at each time $t$, we require the object's state to be always caged such that the object's geometric shape (i.e., the bounding circle), when at this state, is inside a manually set circle $\mathcal{S}_\mathcal{A}^t$.
The center of the circle $\mathcal{S}_\mathcal{A}^t$ is at $(x^c_t, y^c_t)$ with a constant radius $R$.
As such, the ``cage region'' $\mathcal{S}_{cage}^t$ in the object's state space is also a circle that shares the same center with $\mathcal{S}_\mathcal{A}^t$ and has a radius of $R - r$, i.e.,
$\mathcal{S}_{cage}^t = \{(x, y) \in \mathbb{R}^2 \mid (x - x^c_t)^2 + (y - y^c_t)^2\leq (R - r)^2\}$.
The radius of $\mathcal{S}_{cage}^t$, which is $R - r$, is called the size of the cage.

\begin{figure}[t]
   \centering \includegraphics[width=0.45\textwidth]{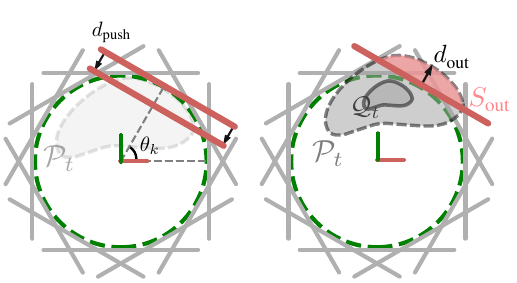}
   \caption{Representation of robot actions and heuristic evaluation for planar pushing. \textit{Left}: The green dashed circle is $\mathcal{S}_\mathcal{A}^t$. Each of the $K=12$ candidate actions (grey bars) is characterized by an angle $\theta_k$. The robot places the line pusher $\mathcal{L}$ (red) at $\theta_k$ and pushes towards the circle center with a distance $d_{push}$. \textit{Right}: Heuristic evaluation, where $S_{out}$ (pink area) is the part of $\mathcal{P}_t$ (the POA) behind the pusher, and $d_{out}$ (black arrow) is the maximum distance to the boundary of $\mathcal{P}_t$.}
   \label{fig:POA11}
   \vspace{-5pt}
\end{figure}

The robot action is characterized by a scalar angle $\theta$, as shown in the left of Fig.~\ref{fig:POA11}.
The robot will first place the center of the line pusher at a starting position determined by $\theta$, which is always on the boundary of the circle $\mathcal{S}_{\mathcal{A}}^t$ and can be calculated by $(x^c_t + R\cos{\theta}, y^c_t + R\sin{\theta})$. 
The line pusher will be oriented to be tangent to the boundary circle of $\mathcal{S}_{\mathcal{A}}^t$.
Then, the robot will move the pusher in the direction towards the center of $\mathcal{S}_\mathcal{A}^t$ with a fixed distance $d_{push}$ to push the object.
We assume there are $K$ candidate robot actions, i.e., $\mathcal{U} = \{\theta_k\}_{k=1}^K$.
The value of each action is obtained through $\theta_k = 2k\pi / K$ such that the starting positions of these $K$ candidate actions will evenly surround $\mathcal{S}_{\mathcal{A}}^t$.

As defined in Sec.~\ref{sec:caging_in_time}, we require the object's state to be always confined inside the cage region $\mathcal{S}_{cage}^t$, i.e., the PSS of the object $\mathcal{Q}_t \subset \mathcal{S}_{cage}^t$.
Then, for the object to follow a certain trajectory $\mathcal{T} = \{(x^d_1, x^d_1), \cdots, (x^d_t, y^d_t), \cdots, (x^d_T, y^d_T)\}$ represented by a sequence of waypoints (i.e., desired positions) in the object's state space, we just need to move the center of $\mathcal{S}_{cage}^t$, which is $(x^c_t, y^c_t)$, along the same trajectory.
As the object is always confined inside the moving $\mathcal{S}_{cage}^t$ for every time step $t = 1, \cdots, T$, the object's position is guaranteed to follow the trajectory $\mathcal{T}$ with a positional error no greater than the cage size $R - r$.

\begin{algorithm}[t]

\SetKwInput{KwData}{Input}
\SetKwInput{KwResult}{Output}
\caption{Pushing by Caging in Time}
\label{alg:pushing}
\KwData{a trajectory as a sequence of waypoints $\mathcal{T} = \{(x^d_1, x^d_1), \cdots, (x^d_t, y^d_t), \cdots, (x^d_T, y^d_T)\}$}
\KwResult{The task finished with success or not}
$\mathcal{Q}_0 \gets \{q_0\}$ \hfill \Comment{Observe the Object's Position}\\
  \For{$t=0, \cdots, T-1$}{
    $(x^c_{t+1}, y^c_{t+1}) \gets (x^d_{t+1}, y^d_{t+1})$ \hfill \Comment{Center of Cage}\\
    $\mathcal{S}_{cage}^{t+1} = \left\{(x, y) \mid (x - x^c_{t+1})^2 + (y - y^c_{t+1})^2 \leq (R - r)^2\right\}$\\
    \hfill \Comment{Construct the Circular Cage}\\
    $u_t \gets \Call{FindPush}{\mathcal{Q}_t, \mathcal{U}, \mathcal{S}_{cage}^{t+1}}$ \hfill \Comment{Alg.~\ref{alg:find_push}}\\
    $\mathcal{Q}_{t+1} \gets \Call{Propagate}{\mathcal{Q}_t, u_t}$ \hfill \Comment{Alg.~\ref{alg:disc_prop}}\\
    \If{$\mathcal{Q}_{t+1} \not\subset \mathcal{S}_{cage}^{t+1}$}{
        \Return{false}\\
    }
  }
  \Return{true}\\
\end{algorithm}

Next, we develop an algorithm for solving the planar pushing problem based on the proposed \emph{Caging in Time} theory.
A high-level description of the developed algorithm is detailed in Alg.~\ref{alg:pushing}.
% The outline of solving the planar-pushing problem with caging-in-time is detailed in Alg.~\ref{alg:pushing}.
At each time step $t$, an open-loop robot action $u_t \in \mathcal{U}$ to cage the object in time will be found, as will be detailed in Sec.~\ref{sec:find_push}; and then the PSS of the object will be propagated according to the robot action $u_t$, which will be discussed in Sec.~\ref{sec:poa_prop}.
If the propagated PSS is always contained in the cage region $\mathcal{S}_{cage}^t$, the object being pushed will follow the desired trajectory along with the moving cage;
otherwise, it is possible that the object is outside the cage, indicating a failure of the task.

\begin{figure}[t]
   \centering \includegraphics[width=0.42\textwidth]{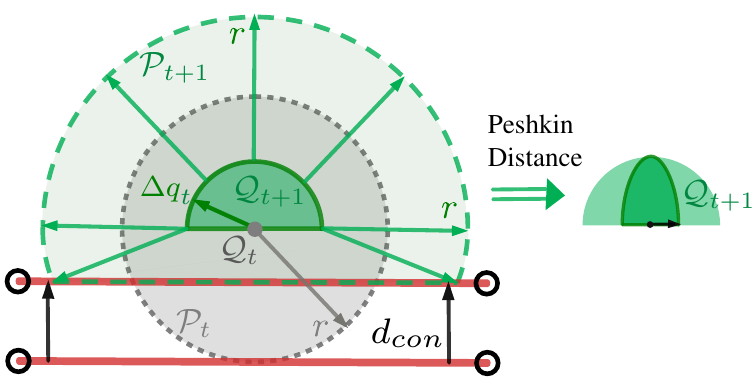}
   \caption{The illustration of the bounds for the object's displacement. The maximum displacement of the object equals $d_{con}$. Therefore, $\mathcal{Q}_t$ containing only the central grey point can propagate into a semi-circle $\mathcal{Q}_{t+1}$ (the semi-circle region in dark green) with a radius equal to $d_{con}$ (i.e., $\lVert \Delta q_t \rVert = d_{con}$). 
   Correspondingly, the POA $\mathcal{P}_t$ (the grey bounding circle) with radius $r$ will evolve into the larger translucent green area $\mathcal{P}_{t+1}$. 
   With analysis based on Peshkin Distance, the possible displacement of the object will be further constrained such that $Q_{t+1}$ becomes a semi-ellipse, as shown on the right side.}
   \label{fig:POA1}
   \vspace{-15pt}
\end{figure}

\subsection{Unknown Object Shape and Its Motion}
\label{sec:poa_prop}
For planar pushing, we represent the object's motion by the displacement of the object's position between adjacent time steps, denoted as $\mathbf{v}_t = \Delta \mathbf{q}_t = (\Delta x_t, \Delta y_t)$.
Then the propagation function $\pi$, initially defined in Sec.~\ref{sec:prop_pcs}, for a single object's configuration and motion as inputs becomes $\pi (\mathbf{q}_t, \mathbf{v}_t) = \pi (\mathbf{q}_t, \Delta \mathbf{q}_t) = \mathbf{q}_t + \Delta \mathbf{q}_t$, which is used to propagate the object's configuration $\mathbf{q}_{t+1}$ at the next step.

For a specific object's configuration $\mathbf{q}_t$, the set of all possible motions $\mathcal{V}_{\mathbf{q}_t}$ is modeled by the interaction between the object and the robot's line pusher $\mathcal{L}$ while the action $u_t$ is being executed, as represented by the function $U$ defined in Eq.~\eqref{eq:U}.
The pusher $\mathcal{L}$ approaches the object with a straight-line distance $d_{push}$ while $u_t$ is being executed.
If the distance between $\mathbf{q}_t$ and the line pusher $\mathbf{d}(\mathbf{q}_t, \mathcal{L})$ is greater than $r + d_{push}$,
the object's bounding circle will not collide with $\mathcal{L}$.
The object will not move at all, and thus $\mathcal{V}_{\mathbf{q}_t} = \left\{ (0, 0) \right\}$.
However, if the object's bounding circle collides with $\mathcal{L}$, indicating the object is likely displaced by $\mathcal{L}$, we denote $d_{con} \leq d_{push}$ as the moving distance of $\mathcal{L}$ after contacting the bounding circle. As shown in Fig. \ref{fig:POA1}, we can derive a bound for this displacement using the friction theory and Peshkin Distance~\citep{peshkin_motion_1988} as follows:

\begin{equation}
\label{eq:Vqt}
    \begin{aligned}
        &\mathcal{V}_{\mathbf{q}_t} = U(\mathbf{q}_t, u_t) = \text{if $\mathbf{d}(\mathbf{q}_t, \mathcal{L}) > r + d_{push}$ or $u_t$ is \textit{None}:}\\
        & \quad \left\{(0, 0)\right\}\\
        &\text{otherwise:}\\
        & \quad \left\{ \mathbf{v}_t = R(u_t) \begin{pmatrix} \Delta x \\ \Delta y \end{pmatrix} \bigg| \frac{\left(\Delta x\right)^2}{d_{con}^2} + \frac{\left(\Delta y\right)^2}{\left(\frac{d_{con}}{2}\right)^2} \leq 1, \Delta x \leq 0 \right\}
        % &\left\{\begin{array}{l}
        % \left\{(0, 0)\right\} \quad \text{if $\mathbf{d}(\mathbf{q}_t, \mathcal{L}) > r + d_{push}$}\\
        % \left\{ \mathbf{v}_t = R(u_t) \begin{pmatrix} \Delta x \\ \Delta y \end{pmatrix} \mid  \frac{\left(\Delta x\right)^2}{d_{con}^2} + \frac{\left(\Delta y\right)^2}{\left(\frac{d_{con}}{2}\right)^2} \leq 1, \Delta x \leq 0\right\}
        % \end{array}\right.
    \end{aligned}
\end{equation}
where $R(u_t) \in SO(2)$ is a rotation matrix of the angle $u_t$. The details of the derivation are given in the Appendix.~\ref{sec:PeskinPush}.

\begin{figure}[t]
   \centering \includegraphics[width=0.5\textwidth]{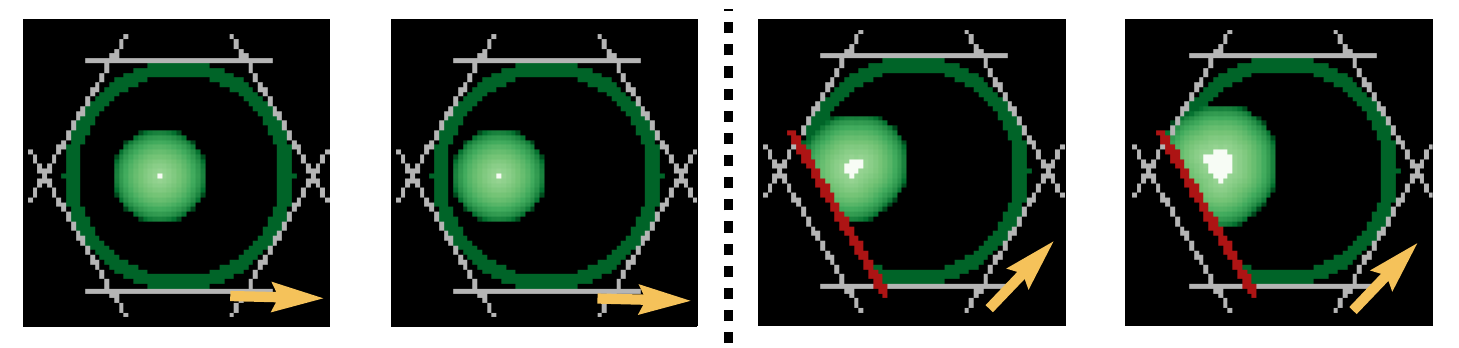}
   \caption{An example showing how the PSS (white region) and the POA (green region) evolve in the discretized space relative to a moving cage. 
   The moving direction of the cage is shown by the yellow arrows.
    \textit{Left}: Without robot intervention, the PSS and the POA remain unchanged. However, they will translate to the left since the cage is moving towards the right. 
    \textit{Right}: If the POA is predicted to be moving outside of $\mathcal{S}_\mathcal{A}^t$ (the dark green circle), a selected line pusher (red) will interact with the POA from the left bottom to deform it to a different shape.}
   \label{fig:POA3}
   \vspace{-20pt}
\end{figure}

The PSS $\mathcal{Q}_t$ and the set of potential motions $\mathcal{V}_{\mathbf{q}_t}$ can both include an infinite number of points in the continuous space, making the propagation of PSS intractable.
To this end, we implement the propagation of PSS in a discretized manner, similar to processing a binary image.
% as exemplified in Fig.~\ref{fig:POA3}.
Specifically, we construct a binary image $\mathcal{I}_t \in \mathbb{R}^{H \times W}$ at each time step $t$ with height $H$ and width $W$, centered at $(x^c_t, y^c_t)$, i.e., the center of the cage.
We use $\mathcal{I}_t(x, y)$ to denote the intensity value of the pixel corresponding to the point $(x, y)$.
An intensity value of $1$ signifies the presence of the point $(x, y)$ in the PSS of the object, and $0$ indicates the absence, i.e., $\mathcal{I}_t(x,y) = 1 \iff (x, y) \in \mathcal{Q}_t$ and $\mathcal{I}_t(x,y) = 0 \iff (x, y) \not\in \mathcal{Q}_t$.
Fig.~\ref{fig:POA3} shows several example images of $\mathcal{P}_t$ enhanced with other features including visuals of the robot actions, the cage, the object's POA, etc.
In the figure, the white pixels (surrounded by the green pixels which are the POA of the object) correspond to the object's PSS, and the pixels in other colors indicate the absence of PSS.

With this discretized representation, the PSS of the object is implemented by a finite set of pixels on $\mathcal{I}_t$.
We also do a similar discretization for the possible motion set $\mathcal{V}_{\mathbf{q}_t}$.
As such, to propagate the PSS on the discretized space, we just enumerate and propagate every pixel in the PSS with every motion in the discretized finite set of $\mathcal{V}_{\mathbf{q}_t}$.
As mentioned, $\mathcal{I}_t$ is always centered with the moving cage.
So we need to deal with an offset of $(x^c_t, y^c_t)$ whenever we convert between a pixel location on $\mathcal{I}_t$ and its actual coordinates in the original continuous space.
Alg.~\ref{alg:disc_prop} outlines the discretized implementation of the PSS propagation.
 \begin{figure*}[t]
   \centering 
   \includegraphics[width=0.96\textwidth]{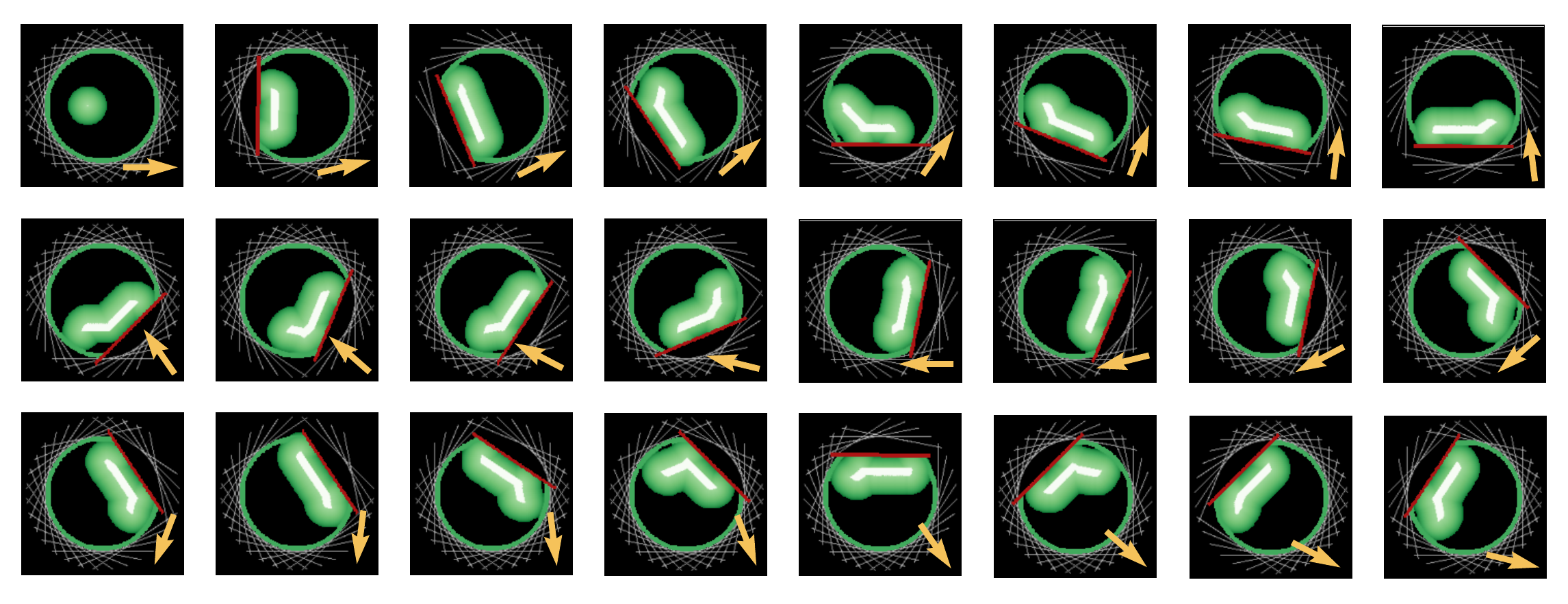}
   \caption{The propagation of PSS (white region) and POA (green region) when the cage is following a circular path. 
   Each image is centered at the center of the cage.
   The yellow arrow in each image represents the motion direction of the cage.
   In this example, $K=32$ candidate actions are used, represented by the grey bars.
   With a different pushing action selected (red bar) for each step, the object is always caged in time, and the POA always stays inside $\mathcal{C}_\mathcal{A}^t$ (the dark green circle).}
   \label{fig:POA}
   \vspace{-10pt}
\end{figure*}
\begin{algorithm}[t]

\SetKwInput{KwData}{Input}
\SetKwInput{KwResult}{Output}
\caption{Propagate($\cdot$)}
\label{alg:disc_prop}
\KwData{The PSS $\mathcal{Q}_t$, a robot action $u_t \in \mathcal{U}$}
\KwResult{The PSS at the next time step $\mathcal{Q}_{t+1}$}
$(x^c_{t+1}, y^c_{t+1}) \gets $ the center of the cage $\mathcal{S}_{cage}^{t+1}$\\
$\mathcal{Q}_{t+1} \gets \{\}$\\
$\mathcal{I}_{t+1} \gets$ initialized with zeros\\
\For{$\mathbf{q}_t \in \mathcal{Q}_t$}{
    $\mathcal{V}_{\mathbf{q}_t} \gets U(\mathbf{q}_t, u_t)$ \hfill \Comment{Eq.~\eqref{eq:Vqt}}\\
    \For{$\mathbf{v}_t = (\Delta x, \Delta y) \in \Call{Discretized}{\mathcal{V}_{\mathbf{q}_t}}$}{
        % \If{$\mathcal{I}_{t+1} (x_t + \Delta x_t - x^c, y_t + \Delta y_t - y^c) == 0$}{
        %     $\mathcal{I}_{t+1} (x_t + \Delta x_t - x^c, y_t + \Delta y_t - y^c) \gets 1$\\
        % }
        $\mathcal{I}_{t+1} (x_t + \Delta x_t - x^c_{t+1}, y_t + \Delta y_t - y^c_{t+1}) \gets 1$\\
        $\mathcal{Q}_{t+1} \gets \mathcal{Q}_{t+1} \cup \{(x_t + \Delta x_t, y_t + \Delta y_t)\}$\\
    }
}
% \For{$(x, y) \in \mathcal{I}_{t+1}$}{
%     \If{$\mathcal{I}_{t+1}(x, y) == 1$}{
%         $\mathcal{Q}_{t+1} \gets \mathcal{Q}_{t+1} \cup \{(x + x^c, y+y^c)\}$
%     }
% }
\Return{$\mathcal{Q}_{t+1}$}

\end{algorithm}

\subsection{Cage the Push in Time}
\label{sec:find_push}

At each time step $t$, we need to select one robot pushing action $u_t$ from all the candidates $\mathcal{U}$.
A naive approach is simulating every action in $\mathcal{U}$, propagating the PSS for every action, and then selecting one of the actions that keep the PSS inside the cage $\mathcal{S}_{cage}^{t+1}$.
However, propagating the PSS can be computationally expensive, especially when the number of candidate actions $K$ is large.
Therefore, we propose a heuristic-based strategy without requiring the PSS propagation for every action being considered, as detailed in Alg.~\ref{alg:find_push}.
It is worth mentioning that the heuristics-based strategy is developed to make the action selection more efficient, alleviating the burden of brute-force search over all possible actions.
It does not conflict with the completeness of the definition of \emph{Caging in Time}.
If the heuristically selected pushing actions are verified to meet the \emph{Caging in Time} condition, the object's state is still guaranteed to be always caged.
However, this work does not focus on developing an optimal strategy for action selection, and we admit a better strategy might be developed in the future.

\begin{algorithm}[t]

\SetKwInput{KwData}{Input}
\SetKwInput{KwResult}{Output}
\caption{FindPush($\cdot$)}
\label{alg:find_push}
\KwData{The PSS $\mathcal{Q}_t$, the set of all candidate robot actions $\mathcal{U}$, the cage region $\mathcal{S}_{cage}^{t+1}$}
\KwResult{The selected robot action $u_t \in \mathcal{U}$ to be taken}
\If{$\mathcal{Q}_t \subset \mathcal{S}_{cage}^{t+1}$}{
    \Return{None}\\

}
\For{$k = 1, \cdots, K$}{

    $h^k \gets \lambda_1 S_{out}^k + \lambda_2 \left(d_{out}^k\right)^2$ \hfill \Comment{Evaluate Heuristics}\\
}
$\mathcal{U}' \gets \{\text{top five $\theta_k$ with greatest $h^k$ }\}$\\
$u_t \gets \arg\min_{\theta \in \mathcal{U'}} \lvert \theta - u_{t-1} \rvert$ \hfill \Comment{Closest to $u_{t-1}$}\\ 
\Return{$u_t$}
\end{algorithm}

As the cage moves from $\mathcal{S}_{cage}^t$ in the previous time step to $\mathcal{S}_{cage}^{t+1}$ along with the desired trajectory, if the PSS is still inside the moving cage, i.e., $\mathcal{Q}_t \subset \mathcal{S}_{cage}^{t+1}$, no robot action will be needed and Alg.~\ref{alg:find_push} will return a \textit{None}.
However, if $\mathcal{Q}_t \not\subset \mathcal{S}_{cage}^{t+1}$, it indicates that the PSS of the object will go outside the cage if no robot action is taken.
In this case, we have to select one robot action to move and deform the PSS to keep the object being caged.

Recall that in Sec.~\ref{sec:prob_stat}, each candidate action in $\mathcal{U}$ is represented by an angle $\theta_k$, which determines the starting pose of the line pusher.
For each candidate action in $\mathcal{U}$ with an index $k = 1, \cdots, K$, we will evaluate a heuristic score by $h^k = \lambda_1 S_{out}^k + \lambda_2 \left(d_{out}^k\right)^2$ where $\lambda_1$ and $\lambda_2$ are weighting factors.
As shown on the right of Fig.~\ref{fig:POA11},
$S_{out}^k$ is the POA area behind the line pusher when we place the pusher at a starting pose determined by the angle $\theta_k$, and $d_{out}^k$ is the furthest distance from behind the pusher to the POA boundary.
It is worth noting that calculating the heuristics $h^k$ for each action does not require the propagation of PSS, i.e., the PSS remains $\mathcal{Q}_t$.
Intuitively, a large $h^k$ indicates that a large portion of the PSS will go outside the cage if the $k$-th action is not taken.

The top five candidate actions with the highest heuristic scores, which compose a set $\mathcal{U}'$, will be further considered for selecting the best one.
From these five candidates in $\mathcal{U}'$, the one closest to the previous action $u_{t-1}$ (i.e., the angle of the last push) will be chosen to propagate the PSS to the next step.
This mechanism reduces the reorientation of the line pusher at adjacent time steps, which can largely facilitate motion efficiency when implemented on a real robot manipulator.
Under the quasi-static setting, we know that the object does not move when not interacting with the robot pusher.
This assumption ensures that the physical robot can switch to the subsequent pushing action in time, without requiring a sufficiently fast speed of operation.
Therefore, the verification of action feasibility (Line $3$ of Alg. \ref{alg:caging_in_time}) is always satisfied. Fig. \ref{fig:POA} demonstrates an overall process of how the PSS and POA evolve as the cage moves, with strategically selected pushing actions ensuring the object remains caged throughout the task.

\vspace{-3pt}
\section{Dynamic Tasks}

In this section, we extend our \textit{Caging in Time} theory from quasi-static tasks to dynamic scenarios where we instantiate the same framework on a challenging dynamic ball balancing problem. Tools are developed for propagating the ball's PSS of its position and velocity, as well as for generating open-loop robot actions to keep the ball balanced while the robot end-effector follows different trajectories without requiring any sensory feedback. This is an instantiation of \emph{Caging in Time} with an energy-based cage.

\subsection{Problem Statement}
\label{sec:dyn_prob_stat}

\begin{figure}[t]
   \centering \includegraphics[width=0.48\textwidth]{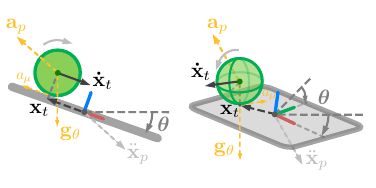}
   \vspace{-10pt}
   \caption{Illustration of the dynamic ball balancing system for plate dimension $n=1$ and $n=2$. \textit{Left}: Physical setup when $n=1$. $\mathbf{x}_t$ and $\dot{\mathbf{x}}_t$ represent the ball's position and velocity respectively. The plate's tilt angle is denoted by $\boldsymbol{\theta}$. Accelerations acting on the ball include gravity $\mathbf{g}_{\theta}$, $\mathbf{a}_p$ induced by the plate's acceleration $\ddot{\mathbf{x}}_p$, and rolling friction $a_\mu$. \textit{Right}: Physical setup when $n=2$, where the notations are the same with $n=1$ case. Note that $\boldsymbol{\theta}$ here is the tilt angle vector that contains two angles.}
   \label{fig:dy_set}
   \vspace{-10pt}
\end{figure}

The robot is tasked with balancing a rolling object on a tilting plate (end-effector) while the plate follows different trajectories in the workspace. The plate is an $n$-dimensional flat surface in a workspace of dimension $n+1$, where $n = 1$ or $2$. Each dimension of the plate has a size ranging from $-l$ to $l$. The object has mass $m$, radius $r_b$, an estimated rolling friction coefficient $\mu_r$, and an estimated moment of inertia $I_b$. We assume that the object only rolls on the plate without slipping.  
Fig. \ref{fig:dy_set} illustrates example setups where $n=1$ and $2$.
The plate's position is denoted as $\mathbf{x}_p \in \mathbb{R}^{n+1}$ and its translational acceleration as $\ddot{\mathbf{x}}_p \in \mathbb{R}^{n+1}$. Its orientation is characterized by a tilt angle vector $\boldsymbol{\theta} \in \mathbb{R}^n$. While only $\boldsymbol{\theta}$ and $\ddot{\mathbf{x}}_p$ are initially defined in the world frame (inertial), all other definitions and models are made in the plate frame (non-inertial). In this case, we introduce the ball's non-inertial acceleration $\mathbf{a}_p \in \mathbb{R}^{n+1}$ and gravity $\mathbf{g}_{\theta} \in \mathbb{R}^{n+1}$ induced by $\ddot{\mathbf{x}}_p$ and $\boldsymbol{\theta}$ so that all the calculations can be done in the plate frame, as shown in Fig. \ref{fig:dy_set}.  

While our framework employs a pure rolling assumption, this approach leverages a physical insight: rolling friction coefficients are typically smaller than sliding, introducing higher dynamic uncertainty. Our \textit{Caging in Time}, specifically designed based on PSS, capitalizes on this property by effectively handling the more uncertain rolling dynamics, inherently encompassing all possible motion states, including sliding behaviors and transitional contact modes, without requiring explicit modeling of each interaction type.

The state of the object at time $t$ is defined as $\mathbf{q}_t = (\mathbf{x}_t, \mathbf{\dot{x}}_t) \in \mathbb{R}^{2n}$, where $\mathbf{x}_t \in \mathbb{R}^n$ is the position vector of the object relative to the plate center, and $\mathbf{\dot{x}}_t \in \mathbb{R}^n$ is its velocity vector. In this dynamic system, the state space is denoted as $\mathcal{S}_{obj} \subset \mathbb{R}^{2n}$, which encompasses all the possible states that the object can have in the plate frame.

To represent the distribution of states in a continuous way, we define the probability density of the object being in state $\mathbf{q}_t$ as $p(\mathbf{q}_t) = p(\mathbf{x}_t, \mathbf{\dot{x}}_t)$. The PSS $\mathcal{Q}_t$ is represented as the set of all possible states with non-zero probability density:
\begin{equation}
\label{eq:Qtdef}
\mathcal{Q}_t = \{\mathbf{q}_t \in \mathcal{S}_{obj} | 
 p(\mathbf{q}_t) > 0\}
\end{equation}
After defining PSS, a ``cage region'' is needed to constrain these states. Unlike quasi-static tasks where the cage is defined geometrically (Sec.~\ref{sec:prob_stat}), dynamic systems require a different approach because their state space encompasses both configuration and velocity. For such systems, energy provides a powerful and intuitive framework for representing and constraining motion-inclusive states. Here we define the cage region in the state space based on system energy:
\begin{equation}
\label{eq:energy}
\mathcal{S}_{cage}^t = \{\mathbf{q}_t \in \mathbb{R}^{2n} | E(\mathbf{q}_t,\mathbf{g}_{\theta}, \mathbf{a}_p) < E_{max}(\mathbf{g}_{\theta}, \mathbf{a}_p)\}
\end{equation}

The total energy of the system $E(\mathbf{q}_t,\mathbf{g}_{\theta}, \mathbf{a}_p)$, for a given state $\mathbf{q}_t= (\mathbf{x}_t, \mathbf{\dot{x}}_t)$ is expressed as:
\begin{equation}
E(\mathbf{q}_t,\mathbf{g}_{\theta}, \mathbf{a}_p) = E_{k}(\mathbf{\dot{x}}_t) + E_{ve}(\mathbf{x}_t) + E_{p}(\mathbf{x}_t, \mathbf{g}_{\theta}, \mathbf{a}_p)
\label{eq:E_xt}
\end{equation}
where $E_{k}$, $E_{ve}$, and $E_{p}$ represent the kinetic energy, virtual elastic energy, and potential energy in the plate frame.

To establish an upper bound on the allowable energy that ensures the object cannot ``escape'' from the plate, we consider an extreme scenario where the object is statically positioned at the edge of the plate as viewed from the plate's frame. This maximum allowable energy, $E_{max}$, is defined as:
\begin{equation}
E_{max}(\mathbf{g}_{\theta}, \mathbf{a}_p) = E_{ve}(l) + E_{p}(l, \mathbf{g}_{\theta}, \mathbf{a}_p)
\label{eq:E_max}
\end{equation}
If the object's total energy exceeds $E_{max}$ at any time step, it would have enough energy to move beyond the plate's boundaries, even if its current position is within the plate. By maintaining the object's energy under the condition $E(\mathbf{q}_t,\mathbf{g}_{\theta}, \mathbf{a}_p) < E_{max}(\mathbf{g}_{\theta}, \mathbf{a}_p)$, we ensure it remains within the plate for all possibilities throughout the task, effectively creating an energy-based ``cage''.

This energy-based definition of the cage region  can capture both position and velocity constraints and establish boundaries that prevent the object from leaving the plate to effectively cage it in the state space and enable robust dynamic object manipulation.

The control action of the robot is defined by adjusting the tilt angle vector $\boldsymbol{\theta}$ of the plate by a continuous variable $\mathbf{u}_t = d\boldsymbol{\theta}(t) \in \mathbb{R}^n$, representing the change rate of the tilt angles. The translation motion of the plate is guided by a predetermined trajectory $\mathcal{T} = \{\mathbf{x}_p(t)\}_{t=0}^T$ in the world frame, where $\mathbf{x}_p(t) \in \mathbb{R}^{n+1}$ represents the position of the plate's center in the workspace. The plate's acceleration trajectory $\mathcal{T}_a = \{\ddot{\mathbf{x}}_p(t)\}_{t=0}^T$ is then calculated using numerical differentiation of the velocity, which in turn is derived from the position trajectory $\mathcal{T}$. According to the theory defined in Sec.~\ref{sec:caging_in_time}, our objective is to maintain the object's state within the energy-constrained cage $\mathcal{S}_{cage}^t$ while the plate follows its prescribed trajectory $\mathcal{T}$, ensuring that the PSS of the object satisfies $\mathcal{Q}_t \subset \mathcal{S}_{cage}^t$ at all times.

Next, we develop an algorithm for solving the dynamic ball balancing problem based on the proposed \emph{Caging in Time} theory. At each time step $t$, we compute the maximum allowable energy, calculate the weighted average energy of the current PSS, and solve an optimization problem to find the optimal control input $\mathbf{u}_t = d\boldsymbol{\theta}(t)$ that keeps the ball caged in time. We then propagate the PSS according to the system dynamics and check if the energy constraint and position constraint are satisfied. If either constraint is violated at any time step, the task is considered failed.

\subsection{Ball Dynamics and PSS Propagation}
\label{sec:dynamic_model}

As aforementioned, the system state at time $t$ is denoted as $\mathbf{q}_t = (\mathbf{x}_t, \dot{\mathbf{x}}_t)$ with its corresponding probability density $p(\mathbf{q}_t)$. We represent the system's motion between adjacent time steps as $\mathbf{v}_{t} = \Delta \mathbf{q}_t = (\dot{\mathbf{x}}_t \Delta t, \ddot{\mathbf{x}}_t \Delta t)$, with associated probability $p(\mathbf{v}_{t})$.  As defined in Eq.~\eqref{eq:U}, the set of potential motions $\mathcal{V}_{\mathbf{q}_t}$ at a certain $\mathbf{q}_t$ can be given as:
\begin{equation}
\label{eq:Vq_d}
\mathcal{V}_{\mathbf{q}_t} = U(\mathbf{q}_t, u_t) = \left\{\mathbf{v}_t\mid \mathbf{v}_t = (\dot{\mathbf{x}}_t \Delta t, \ddot{\mathbf{x}}_t \Delta t) \right\}
\end{equation}

The propagation function $\pi$ that propagate from $\mathbf{q}_{t}$ to $\mathbf{q}_{t+1}$ for a single state-motion pair $(\mathbf{q}_{t}, \mathbf{v}_{t})$ is given with the joint probability $p(\mathbf{q}_{t}, \mathbf{v}_{t})$ by:
\begin{equation}
\pi(\mathbf{q}_{t}, \mathbf{v}_{t}) = (\mathbf{x}_{t} + \dot{\mathbf{x}}_t \Delta t, \dot{\mathbf{x}}_{t} + \ddot{\mathbf{x}}_t \Delta t)
\end{equation}
Note that $\pi$ is a deterministic function for state transition. Unlike Sec. \ref{sec:poa_prop}, here we propagate the PSS by updating its probability distribution. Since the joint probability of the state-motion pair can be expressed as $p(\mathbf{q}_{t}, \mathbf{v}_{t}) = p(\mathbf{q}_{t}) \cdot p(\mathbf{v}_{t}|\mathbf{q}_{t}) $, the probability $p(\mathbf{q}_{t+1})$ can then be derived as:
\begin{equation}
\label{eq:probqp}
\begin{aligned}
p(\mathbf{q}_{t+1}) &= \iint\limits_{\mathbf{q}_{t} \in \mathcal{Q}_t,\mathbf{v}_{t} \in \mathcal{V}_{\mathbf{q}_t}} p(\mathbf{q}_{t},\mathbf{v}_{t}) \mathbbm{1}(\pi(\mathbf{q}_{t}, \mathbf{v}_{t}) =\mathbf{q}_{t+1}) d\mathbf{v}_{t} d\mathbf{q}_{t}\\
&= \int_{\mathbf{q}_{t} \in \mathcal{Q}_t} p(\mathbf{q}_{t}) \int_{\mathbf{v}_{t} \in \mathcal{V}_{\mathbf{q}_t}} \\
&\quad\quad \quad  p(\mathbf{v}_{t}|\mathbf{q}_{t})\mathbbm{1}(\pi(\mathbf{q}_{t}, \mathbf{v}_{t}) =\mathbf{q}_{t+1}) d\mathbf{v}_{t} d\mathbf{q}_{t}
\end{aligned}
\end{equation}
where the decomposition of $p(\mathbf{v}_{t}|\mathbf{q}_{t})$ can be given as:
\begin{equation}
\label{eq:pqv}
\begin{aligned}
p(\mathbf{v}_{t}|\mathbf{q}_{t}) &= p(\dot{\mathbf{x}}_t|\mathbf{q}_{t}) \cdot p(\ddot{\mathbf{x}}_t|\mathbf{q}_{t}) \\
                                  &= p(\ddot{\mathbf{x}}_t|\mathbf{q}_{t}) \\
                                  &= p(\ddot{\mathbf{x}}_t)
\end{aligned}
\end{equation}
where $p(\dot{\mathbf{x}}_t|\mathbf{q}_{t}) = 1$ since $\dot{\mathbf{x}}_t$ is fully determined by $\mathbf{q}_{t}$. To calculate $p(\ddot{\mathbf{x}}_t)$, we need to consider the system dynamics. Based on the physical properties of the ball mentioned in Sec.~\ref{sec:dyn_prob_stat}, we can define the set of potential accelerations $\mathcal{A}_{\mathbf{q}_t}$ for a given state $\mathbf{q}_t$ and control input $u_t$ as:
\begin{equation}
\begin{aligned}
\mathcal{A}_{\mathbf{q}_t} = \Big\{ \ddot{\mathbf{x}}_t \mid &\ddot{\mathbf{x}}_t = f(\mathbf{q}_t, u_t, \eta_m, \boldsymbol{\eta}_p, \eta_\mu), \\
    &\eta_m \sim \mathcal{N}(0, \sigma^2_m), \\
    &\boldsymbol{\eta}_p \sim \mathcal{N}(\mathbf{0}, \Sigma_p), \\
    &\eta_\mu \sim \mathcal{N}(0, \sigma^2_\mu) \Big\}
\end{aligned}
\end{equation}
where $f(\cdot)$ represents the system dynamics equation, detailed in the Appendix~\ref{sec:appendix_dynamics}. The terms $\eta_m$, $\boldsymbol{\eta}_p$, and $\eta_\mu$ represent uncertainties in mass, plate acceleration, and friction coefficient, respectively.

Leveraging the closure property of Gaussian distributions under linear operations, we can conclude that the acceleration $\ddot{\mathbf{x}}_t$, being a linear combination of Gaussian-distributed uncertainties, also follows a Gaussian distribution:
\begin{equation}
\ddot{\mathbf{x}}_t \sim \mathcal{N}(\boldsymbol{\mu}_{\ddot{\mathbf{x}}}, \boldsymbol{\Sigma}_{\ddot{\mathbf{x}}})
\end{equation}
where $\boldsymbol{\mu}_{\ddot{\mathbf{x}}}$ and $\boldsymbol{\Sigma}_{\ddot{\mathbf{x}}}$ are derived from the system dynamics equation and the distributions of the uncertainty terms.

Given this Gaussian distribution of $\ddot{\mathbf{x}}_t$, we can now compute $p(\ddot{\mathbf{x}}_t)$ for any $\ddot{\mathbf{x}}_t \in \mathcal{A}_{\mathbf{q}_t}$, which allows us to further calculate $p(\mathbf{q}_{t+1})$ based on Eq.~\eqref{eq:probqp} and Eq.~\eqref{eq:pqv}.

As the probability of $p(\mathbf{q}_{t+1})$ is propagated from all the states in $\mathcal{Q}_t$ , we can get the PSS at next time step in the same way as Eq.~\eqref{eq:Qtdef}: 
\begin{equation}
\label{eq:Qt1def}
\mathcal{Q}_{t+1} = \{\mathbf{q}_{t+1} \in \mathcal{S}_{obj} | 
 p(\mathbf{q}_{t+1}) > 0\}
\end{equation}

\begin{algorithm}[t]

\SetKwInput{KwData}{Input}
\SetKwInput{KwResult}{Output}
\caption{DynamicPropagate($\cdot$)}
\label{alg:dynamic_prop_simplified}
\KwData{Current PSS $\mathcal{Q}_t$, a robot action $u_t\in\mathcal{U}$, plate acceleration $\ddot{\mathbf{x}}_p$, time step $\Delta t$}
\KwResult{The PSS at next time step $\mathcal{Q}_{t+1}$}
$\mathcal{Q}_{t+1} \gets \{\}$\\
$\mathcal{I}_{t+1} \gets$ initialized with zeros\\
\For{$q_t \in \mathcal{Q}_t$}{
$\mathcal{V}_{\mathbf{q}_t} \gets U(\mathbf{q}_t, u_t)$ \hfill \Comment{Eq.~\eqref{eq:Vq_d}}\\
\For{$\mathbf{v}_t = (\dot{\mathbf{x}}_t \Delta t, \ddot{\mathbf{x}}_t \Delta t) \in \Call{Discretized}{\mathcal{V}_{\mathbf{q}_t}}$}{

$\mathbf{q}_{t+1} \gets \Call{Discretized}{\pi(\mathbf{q}_{t}, \mathbf{v}_{t})}$\\
\If{$\mathbf{q}_{t+1}\in\mathcal{S}_{obj}$}{
$P_{q,v} \gets \Call{CalcProb}{\mathbf{q}_t,\mathbf{v}_t}$\hfill \Comment{Eq.~\eqref{eq:pqv}}\\

$\mathcal{I}_{t+1}(\mathbf{q}_{t+1}) \gets \mathcal{I}_{t+1}(\mathbf{q}_{t+1})+ P_{q,v}$\
}
}
}
$\mathcal{I}_{t+1} \gets \Call{Normalize}{\mathcal{I}_{t+1}}$\\
\Return{$\mathcal{I}_{t+1}$}
\end{algorithm}

\begin{figure}[t]
   \centering \includegraphics[width=0.3\textwidth]{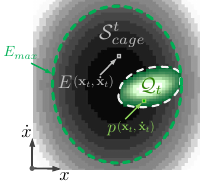}
   \caption{Discretized representation of the state space $\mathcal{S}_{obj}$ with energy map. Each pixel of the discretized map represent a state $(\mathbf{x}_t,\dot{\mathbf{x}}_t)$ of the ball. The cage region $\mathcal{S}_{cage}^t$ is bounded by the maximum energy $E_{max}$. The gradient gray area shows energy value $E(\mathbf{x}_t,\dot{\mathbf{x}}_t)$ at different states. And the gradient green area shows PSS $\mathcal{Q}_t$ with the ball's probability density $p(\mathbf{x}_t,\dot{\mathbf{x}}_t)$ at different states.}
   \label{fig:PSS}
   \vspace{-10pt}
\end{figure}

To numerically implement the PSS propagation described by Eq. \eqref{eq:Qt1def} and Eq. \eqref{eq:probqp}, we employ a discretized representation of the state space, similar to the approach used in the quasi-static case (Sec.~\ref{sec:poa_prop}). However, instead of binary values, we use continuous values to represent the probability distribution of the PSS in the state space.

We construct a 2n-dimensional array $\mathcal{I}_t \in \mathbb{R}^{N ^ {2n}}$ at each time step $t$, where $N$ is the grid size for each dimension of the state. The array $\mathcal{I}_t$ is centered at the origin in the state space, with $\mathbf{x} \in [-\mathbf{x}_{max}, \mathbf{x}_{max}]$ and $\dot{\mathbf{x}} \in [-\dot{\mathbf{x}}_{max}, \dot{\mathbf{x}}_{max}]$. 
% Each element $\mathcal{I}_t(\mathbf{i})$ represents the probability value of the grid point corresponding to the state $(\mathbf{x}_t, \dot{\mathbf{x}}_t)$, where $\mathbf{i} = [i_1, i_2, ..., i_{2n}]^T \in \mathbb{N}^{2n}$. 
The state space is discretized uniformly in each dimension, with appropriate grid spacings for position and velocity components.
Each element $\mathcal{I}_t(\mathbf{q}_t)$ represents the probability of the occurrence of the discretized version of state $\mathbf{q}_t$, derived from $p(\mathbf{q}_t)$.
Fig. \ref{fig:PSS} shows a visual illustration under the above numerical representation when $n=1$.

The PSS propagation is implemented in Alg. \ref{alg:dynamic_prop_simplified}, which discretizes the continuous integrals in Eq. \eqref{eq:probqp}. For each $\mathbf{q}_t$ in the current PSS $\mathcal{Q}_t$, we compute the set of potential motions $\mathcal{V}_{\mathbf{q}_t}$ using Eq. \eqref{eq:Vq_d}. We then iterate over discretized versions of these motions, updating the probability of the resulting states $\mathbf{q}_{t+1}$ according to Eq. \eqref{eq:pqv}:
\begin{equation}
\mathcal{I}_{t+1}(\mathbf{q}_{t+1}) = \mathcal{I}_{t+1}(\mathbf{q}_{t+1}) + \mathcal{I}_t(\mathbf{q}_t) \cdot p(\mathbf{v}_t|\mathbf{q}_t)
\end{equation}
where $p(\mathbf{v}_t|\mathbf{q}_t)$ is computed as in Eq. \eqref{eq:pqv}. After updating all probabilities, we normalize it to ensure the values of the grid sum to 1 as a valid distribution. For computational efficiency, we apply a threshold to $\mathcal{I}_{t+1}$, setting probabilities below a certain value (e.g., $10^{-3}$) to zero.
\begin{figure}[t]
   \centering \includegraphics[width=0.48\textwidth]{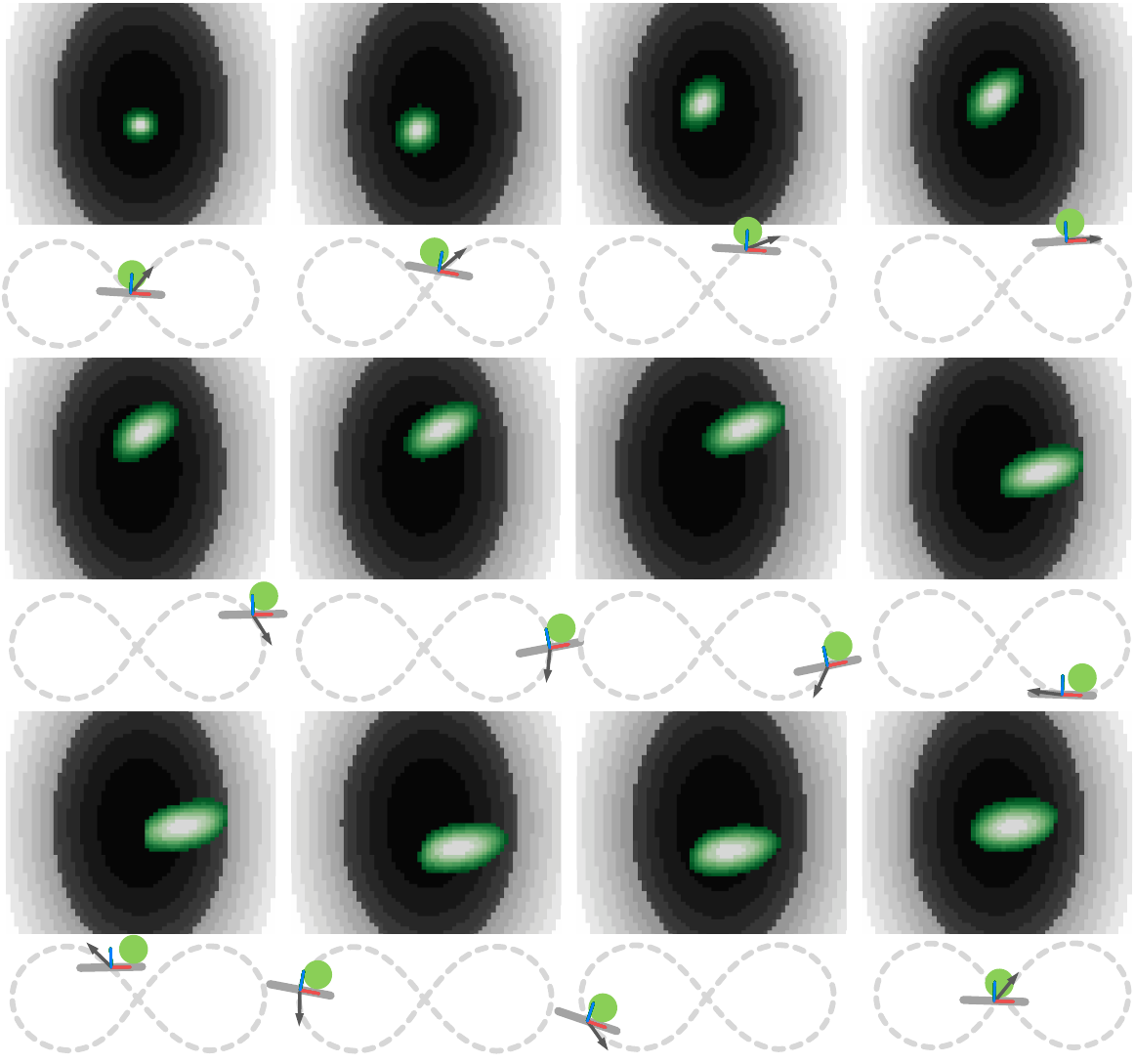}
   \caption{Sequence of the ball's PSS evolution during dynamic balancing where the plate dimension $n=1$. \textit{Top rows}: State space representation of the PSS $\mathcal{Q}_t$ (green) within the cage region $\mathcal{S}_{cage}^t$ (dark background), as detailed in Fig.~\ref{fig:PSS}. \textit{Bottom rows}: Corresponding configurations of the system in the world frame. The dark gray arrow shows the direction of the plate's translational motion.} 
   \label{fig:dynamPSS}
   \vspace{-15pt}
\end{figure}
This pixel-based approach provides a computationally efficient representation of the PSS in the state space, avoiding the exponential increase in computational cost associated with sampling methods used in previous caging work in long-horizon tasks \citep{welle_partial_2021,makapunyo_measurement_2012}.
% The continuous-valued representation allows for an accurate model of the probability distribution of the ball's state over time, accounting for the uncertainties in the system dynamics. 
Fig.~\ref{fig:dynamPSS} illustrates the propagation and caging of the PSS in the state space during the tracing of an $\infty$-shaped trajectory for a one-dimensional plate ($n=1$). The PSS initially manifests as a compact Gaussian distribution, subsequently expanding before converging to a stable size caged by the energy boundary.

\subsection{Cage the Ball in Time}

To achieve \textit{Caging in Time} for the dynamic ball balancing, we employ a control optimization strategy that combines Control Barrier Function \citep{aaron2019cbf} and Control Lyapunov Function \citep{ANAND20213987} to keep the ball on the plate while following the desired trajectory.

In our case, Control Barrier Function (CBF) is utilized to implement the concept of ``caging'' in dynamic setting. As a physical cage constraining an object within a defined space, the CBF establishes an energy-based cage that can be characterized by the margin between the maximum energy attainable within the PSS and the preset energy boundary:
\begin{equation}
\label{eq:cbf}
h(\mathcal{Q}_t) = E_{max}(\mathbf{g}_{\theta}, \mathbf{a}_p) - \max_{\mathbf{q}_t\in \mathcal{Q}_t} E(\mathbf{q}_t,\mathbf{g}_{\theta}, \mathbf{a}_p)
\end{equation}

This creates an energy barrier and acts as the mathematical representation of our cage. By ensuring that $h(\mathcal{Q}_t) > 0$ at all times, we guarantee that the ball's energy, $E(\mathbf{q}_t,\mathbf{g}_{\theta}, \mathbf{a}_p)$ at state $\mathbf{q}_t$ as calculated in Eq.~\eqref{eq:E_xt}, never exceeds the maximum allowable energy $E_{max}$ defined in Eq.~\eqref{eq:E_max}. 

To further ensure our control actions maintain this energy barrier, we enforce a CBF condition. This provides a forward-looking constraint that considers not only the current state but also the system's future behavior, rather than simply enforcing $h(\mathcal{Q}_t) > 0$.

\begin{equation}
\label{eq:cbf_condition}
L_f h(\mathcal{Q}_t) + L_g h(\mathcal{Q}_t)d\boldsymbol{\theta} + \alpha (h(\mathcal{Q}_t)) \geq 0
\end{equation}

In this condition, $L_f h(\mathcal{Q}_t)$ and $L_g h(\mathcal{Q}_t)$ are the Lie derivatives of the CBF $h(\mathcal{Q}_t)$ along the vector fields $f$ and $g$ respectively, where $f$ and $g$ are defined in the system dynamics equation $\ddot{\mathbf{x}}_t = f + g u$, detailed in Appendix~\ref{sec:cbfclf}. $L_f h(\mathcal{Q}_t)$ represents the change in the barrier function due to the natural dynamics of the system, while $L_g h(\mathcal{Q}_t)d\boldsymbol{\theta}$ represents the change due to the control input. The term $\alpha(\cdot)$ is a class $\mathcal{K}$ function that shapes the convergence behavior of the barrier function, ensuring that it increases more rapidly as the system approaches the boundary of the safe set. 
% This condition is more powerful than simply enforcing $h(\mathcal{Q}_t) > 0$ as it provides a forward-looking constraint that considers not just the current state but also the system's future behavior. 

Control Lyapunov Functions (CLFs), on the other hand, are used to optimize the control performance by driving the system towards a desired state. In our context, we use a CLF to encourage the ball to stay near the center of the plate and maintain a well-distributed probability state, with both the system's energy and entropy considered:
\begin{equation}
\label{eq:clf}
V(\mathcal{Q}_t) = \sum_{\mathcal{I}_t(\mathbf{q}_t) > 0} \mathcal{I}_t(\mathbf{q}_t)E(\mathbf{q}_t,\mathbf{g}_{\theta}, \mathbf{a}_p) - k_S S(\mathcal{Q}_t)
\end{equation}
where $\mathcal{I}_t(\mathbf{q}_t)$ is the probability value at state $\mathbf{q}_t$ in the discretized probability distribution, $E(\mathbf{q}_t,\mathbf{g}_{\theta}, \mathbf{a}_p)$ is the system energy at the state $\mathbf{q}_t$, $k_S$ is a weighting factor, and $S(\mathcal{Q}_t)$ is the system entropy. 

% And maintaining a well-distributed probability state, ensuring both stability and robustness in the face of uncertainties.

This CLF balances two objectives: minimizing the expected energy of the system, which tends to keep the ball near the center of the plate, and maximizing the entropy of the probability distribution, which maintains a well-distributed probability state, ensuring both stability and robustness in the face of uncertainties. By minimizing this CLF, we drive the system towards a state where the ball is likely to be near the center of the plate, but with some uncertainty to handle unexpected disturbances.

Similar to CBF condition, we enforce the CLF condition to ensure the CLF decreases over time:
\begin{equation}
\label{eq:clf_condition}
L_f V(\mathcal{Q}_t) + L_g V(\mathcal{Q}_t)d\boldsymbol{\theta} + cV(\mathcal{Q}_t) \leq \delta
\end{equation}
where $L_f V(\mathcal{Q}_t)$ and $L_g V(\mathcal{Q}_t)$ are Lie derivatives of the CLF along the system dynamics and control input respectively, $c$ is the convergence rate, and $\delta$ is a relaxation variable. This condition ensures that our control input $d\boldsymbol{\theta}$ drives the system towards the desired behavior at a sufficient rate, while the relaxation variable $\delta$ allows for temporary violations of the decrease condition when necessary to satisfy the CBF safety constraint.

By combining the CBF and CLF conditions, we formulate a Quadratic Program (QP) that not only keeps the ball on the plate but also tries to keep it centered and stable.
\begin{equation}
\label{eq:qp}
\begin{aligned}
\min_{d\boldsymbol{\theta}, \delta} \quad & (d\boldsymbol{\theta})^2 +  \lambda\delta^2 \\
\text{s.t.} \quad & L_f h(\mathcal{Q}_t) + L_g h(\mathcal{Q}_t)d\boldsymbol{\theta} + \alpha(h(\mathcal{Q}_t)) \geq 0 \\
& L_f V(\mathcal{Q}_t) + L_g V(\mathcal{Q}_t)d\boldsymbol{\theta} + c(V(\mathcal{Q}_t)) \leq \delta \\
& d\boldsymbol{\theta}_{\min} \leq d\boldsymbol{\theta} \leq d\boldsymbol{\theta}_{\max}
\end{aligned}
\end{equation}
Here, $\lambda > 0$ is a weighting factor that balances the trade-off between minimizing the control effort $(d\boldsymbol{\theta})^2$ and the CLF constraint violation $\delta^2$. The values of $d\boldsymbol{\theta}_{\min}$ and $d\boldsymbol{\theta}_{\max}$ are derived from the hardware torque limits of the robot arm's motors. We utilize the OSQP (Operator Splitting Quadratic Program) solver to efficiently solve this optimization problem at each time step.

By solving this QP at each time step, we obtain the optimal control input that maintains the ball within the energy-based cage while following the desired trajectory. The detailed implementation is presented in the Appendix~\ref{sec:cbfclf}.

This approach allows us to cage the ball in time under dynamic conditions on a tilting plate, accounting for the ball's dynamics, uncertainties, and energy constraints. By solving the QP at each time step, we generate a sequence of actions that keeps the ball within the energy-based cage region with the plate following the desired trajectory.
\begin{figure}[t]
   \centering \includegraphics[width=0.48\textwidth]{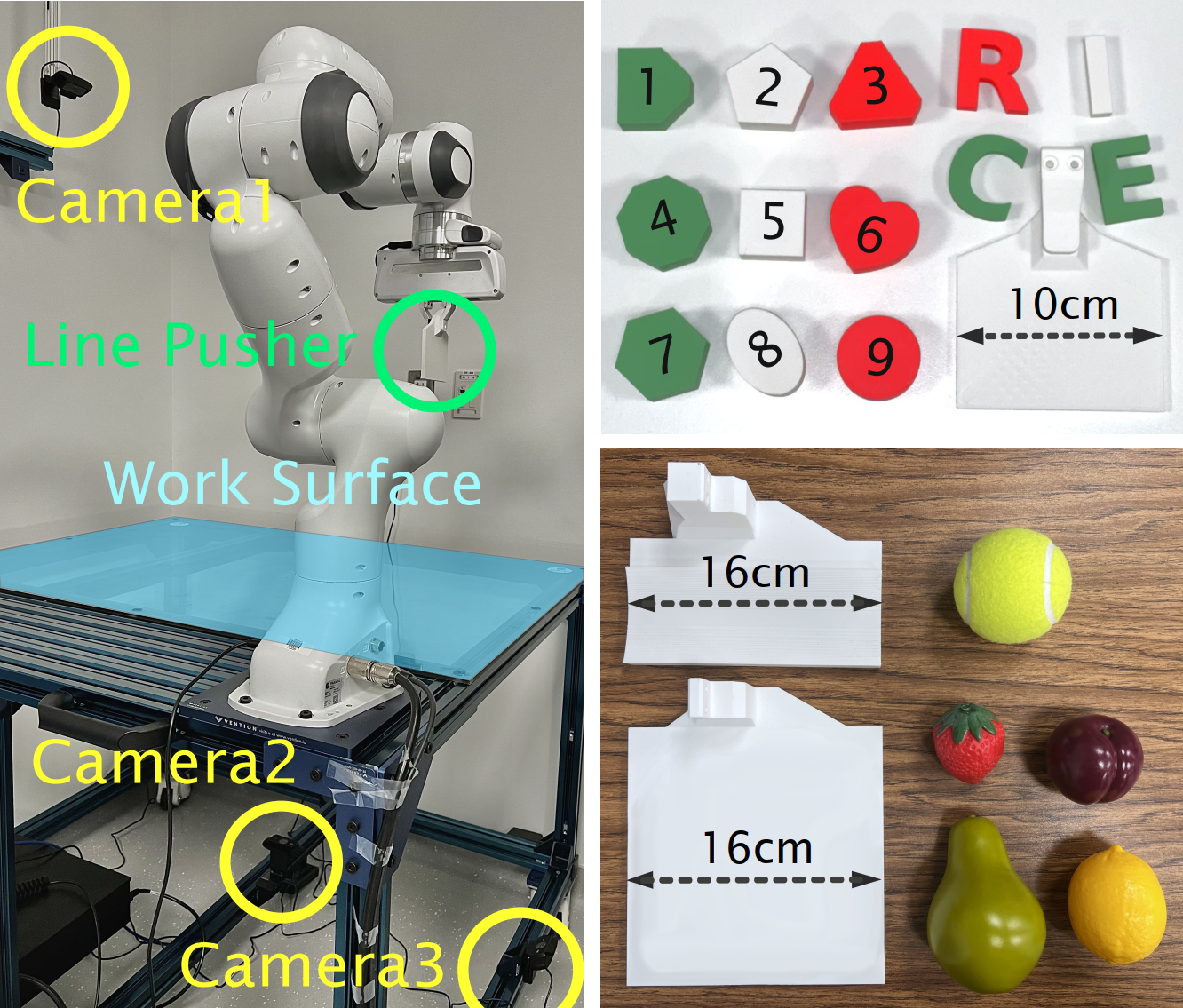}
   \caption{Experiment setup. \textit{Left}: Robot and camera setup. The robot performs the experiment on a transparent surface (blue) with a pre-installed line pusher (green). We used cameras 2 and 3 below the surface to record (not to track) the trajectories of the objects through April-Tag. Camera 1 was used to record the experiment from an upper view. \textit{Top Right}: The 3D-printed objects used in the experiments. 1. Square with two trimmed tips, 2. Pentagon,  3. Triangle with trimmed tips, 4. Octagon, 5. Square, 6. Heart, 7. Hexagon, 8. Ellipse, 9. Circle, and letter-shaped objects: R, I, C, E. \textit{Bottom Right}: 3D printed plates, tennis ball and plastic fruits for dynamic tasks. The square plate is for ball balancing when dimension $n=2$, and the thinner one is for ball balancing and catching when dimension $n=1$, which has a guide rail to constrain the motion of the ball only on the required dimension. The plastic fruits are strawberry (YCB \#12), plum (YCB \#18), pear (YCB \#16) and lemon (YCB \#14).}

   \label{fig:setup}
   \vspace{-10pt}
\end{figure}

\section{Experiments}
\label{sec:experiments}

We conducted extensive experiments to evaluate our \textit{Caging in Time} framework using a Franka Emika Panda robot arm for both quasi-static pushing tasks and dynamic ball balancing tasks with different end-effectors shown in Fig.~\ref{fig:setup}. All algorithms were implemented in Python and executed on a single thread of a 3.4 GHz AMD Ryzen 9 5950X CPU.

Our experimental setup utilized multiple cameras for recording: cameras shown on the left of Fig.~\ref{fig:setup} for quasi-static tasks, with an additional camera for dynamic tasks. For quasi-static experiments, object tracking was achieved using AprilTags \citep{olson2011apriltag}, while in dynamic scenarios, OpenCV was employed to track the ball's motion. It's important to note that object tracking served only for trajectory visualization and accuracy evaluation and all experiments were conducted in an open-loop manner.

To quantify manipulation precision, we adopted the Mean Absolute Error (MAE) as our primary metric, calculated by averaging the absolute distances between the actual and reference trajectories throughout the manipulation process. 

As this initial exposition of our framework, we deliberately selected two representative manipulation paradigms — one quasi-static and one dynamic — to evaluate fundamental capabilities while facilitating clear visualization of PSS propagation. Through these experiments, we aimed to thoroughly assess the performance, robustness, and versatility of our \textit{Caging in Time} framework, establishing a foundation for future extensions to more scenarios.

\subsection{Quasi-static Tasks}
 Through quasi-static experiments, we aimed to investigate the performance of our framework from three aspects: 1) Without any sensing feedback besides the size and initial position of the bounding circle, can robust planar manipulation be achieved by the proposed \textit{Caging in Time}? 2) If so, what precision can be achieved, and how is the robustness affected by different settings of the framework and outer disturbances? 3) Under imperfect perceptions like positional noise and network lag, can \textit{Caging in Time} surpass closed-loop methods?

As displayed in Fig. \ref{fig:setup}, the robot in this task used a 100 mm line pusher, and objects of various shapes were 3D-printed in a similar size, fitting within a bounding circle of radius $r = 25$ mm. The pushing distance of each action was set as $d_{push}=20$ mm. The computation time for each step in action selection and PSS propagation was $26.2 \pm 8.5 $ ms.

\subsubsection{Evaluation of Cage Settings}

\begin{figure}[t]
   \centering \includegraphics[width=0.44 
  \textwidth]{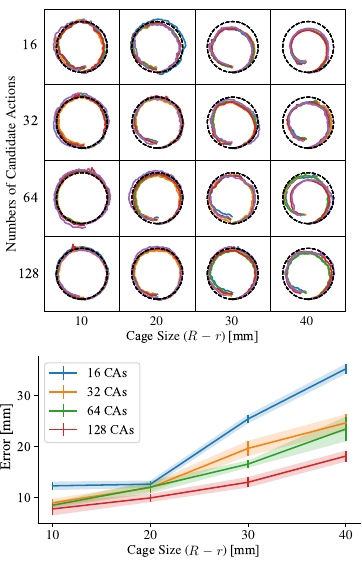}
   \caption{Evaluation results in terms of the cage size ($R-r$) and number of candidate actions. \textit{Top}: All trajectories (nine trajectories of different objects are overlaid in the plot per circle) \textit{Bottom}: Mean Absolute Error of the shown trajectories under different cage settings. ``CA'' stands for candidate action.}
   \label{fig:quant}
\end{figure}

The task of this experiment is to push the object through a predefined circular trajectory. This experiment focused on evaluating the precision of our framework by varying two key parameters: $R-r$, the cage size, and $K$, the number of candidate actions. We selected values for $R-r$ and $K$ from the sets $R-r = \{10, 20, 30, 40\}$ mm and $K = \{16, 32, 64, 128\}$. For each combination of $R-r$ and $K$, five trials were conducted using five different objects (Object 1-5, see Fig.~\ref{fig:setup}). We generated only one robot action sequence per $(R-r, K)$ setting, following Alg. \ref{alg:pushing} and applied it to different objects in an open-loop manner. 
% Fig. \ref{fig:POA} shows an example of how we generate the action sequence with PSS (and POA) propagation where $K=32$. 
The MAE across these trials was averaged and presented in Fig.~\ref{fig:quant}, which also includes the actual trajectories recorded from all experiments with different settings and different objects.
\begin{figure}[t]
   \centering \includegraphics[width=0.5\textwidth]{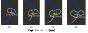}
   \caption{Visualization of POAs (the white boundaries) in real pushing experiment recordings under various cage sizes.}
   \label{fig:poaexp}
   \vspace{-15pt}
\end{figure}

From the results, we can see that the performance follows two major trends. One is that a larger cage radius tends to increase the average error. The other trend is that an increase in the number of candidate actions generally leads to a decrease in error. As shown in Fig. \ref{fig:poaexp}, with larger cage sizes, the POA will grow larger, hence increasing the range the object could possibly deviate from the reference trajectory. Also with more candidate actions, it increases the possibility of finding the optimal action in Alg. \ref{alg:find_push}.

Examining the trajectories in Fig. \ref{fig:quant}, we observed that all paths remained within their respective cage regions. Notably, trajectories generated by the same action sequence showed similar patterns across different objects. For example, trajectories with 16 candidate actions and a 40 mm cage size ($R-r$) mostly moved to the left of the reference path, indicating that the action sequence plays a more dominant role in the performance than the object shape.
\begin{figure}[t]
   \centering \includegraphics[width=0.5\textwidth]{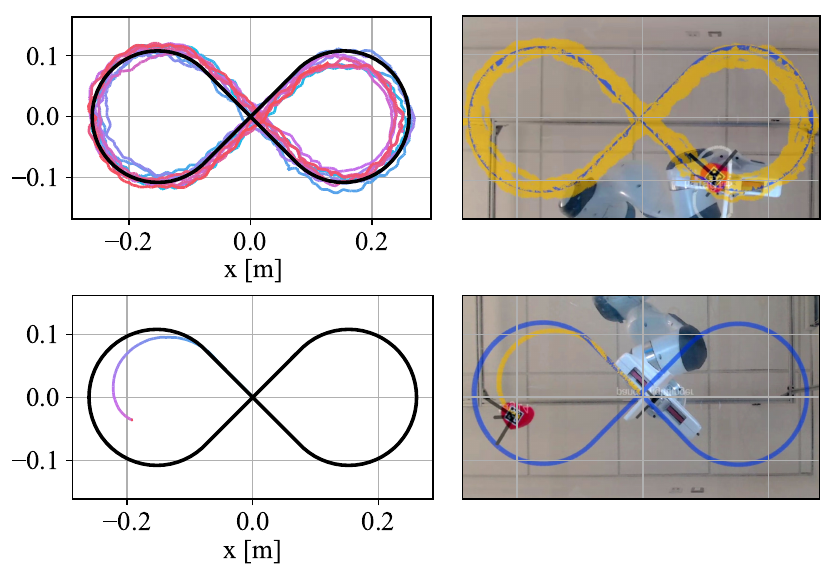}
   \caption{Robustness validation with a $\infty$-shaped trajectory. This experiment was conducted with the heart-shaped object (9), recorded by camera 2. \textit{Top}: Ten loops conducted by \textit{Caging in Time}. \textit{Bottom}: A failed attempt by a na\"ive pushing strategy. }
   \label{fig:robust}
   \vspace{-10pt}
\end{figure}

\begin{figure}[t]
\centering
\includegraphics[width=0.48\textwidth]{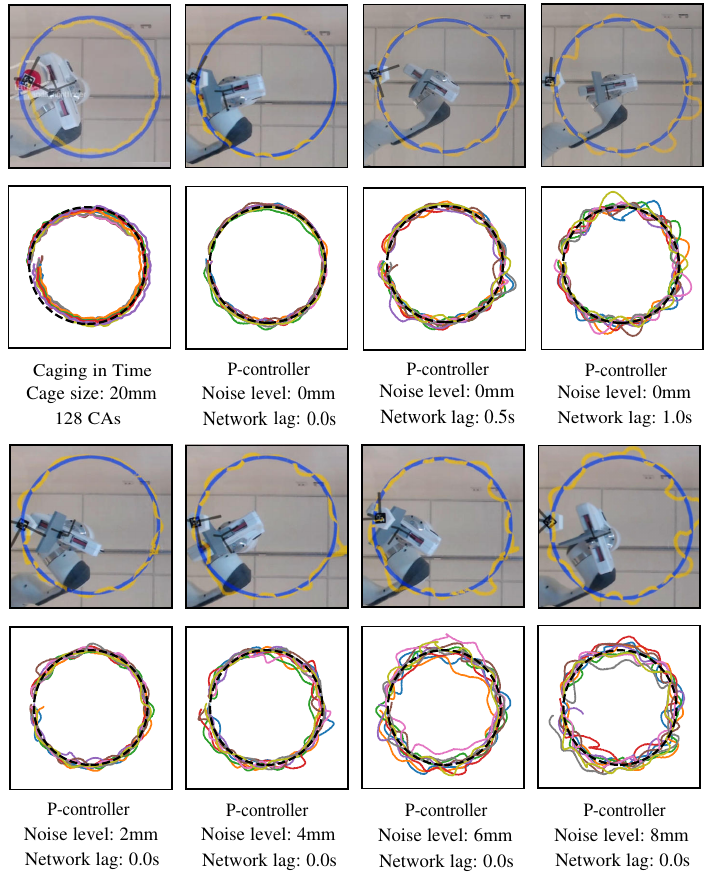}
\caption{Trajectory comparison between \textit{Caging in Time} with $20$mm cage size and 128 candidate actions and the P-controller under varying positional noise levels and network lags. \textit{Odd rows}: Representative recordings of a single object. \textit{Even rows}: Overall trajectories from all nine test objects.}
\label{fig:gauss}
\vspace{-15pt}
\end{figure}

% \begin{figure}[t]
%    \centering \includegraphics[width=0.5\textwidth]{figs/lag_.pdf}
%    \caption{Trajectory comparison between \textit{Caging in Time} with $20$mm cage size and 128 candidate actions and the P-controller under different network lags. \textit{Odd rows}: The actual recording of a specific object. \textit{Even rows}: Overall performance across all nine objects.}
%    \label{fig:lag}
% \end{figure}

% Next, based on the analysis above, we further evaluate the performance of trajectory tracking by setting a cage size of 20 mm and 128 candidate actions for all subsequent experiments. We utilized all nine objects besides ``RICE'' in Fig. \ref{fig:setup}. For the circular trajectory, the average tracking error across all nine objects was  $12.95\pm1.61$ mm.

\subsubsection{Why Caging in Time}

% why caging is necessary

%people may think line pusher is stable, but it's not cuz uncertianty. openloop different strategy, planning, as many times as you need. story

While the line pusher might be perceived as a stable method for pushing an object in an open-loop way, this stability is often compromised by the unpredictability of the object's physical properties and shape, which necessitates the implementation of the \emph{Caging in Time}.

To validate the necessity and robustness of our approach, an object was tasked to navigate a $\infty$-shaped trajectory repeatedly for 10 cycles. A comparative analysis was conducted against a na\"ive pushing strategy, where the pusher simply followed the target trajectory with its orientation parallel to the trajectory direction. As shown in Fig. \ref{fig:robust}, in contrast to the na\"ive method, where the pusher lost control of the object after a few steps, our approach successfully maintained control over the object after 10 rounds with an average error of $10.09$ mm.

\subsubsection{Comparison with a Closed-loop Method}

To further validate the effectiveness of our framework, we conducted a comparative study against a closed-loop pushing method: a proportional controller (P-controller) implemented under the quasi-static assumption. The P-controller works by calculating an error vector from the object's current position to the nearest waypoint on the reference trajectory. This error vector is then scaled by the P-gain (set to 0.5) to determine the direction and magnitude of the pushing action. Specifically, the pushing direction is aligned with the scaled error vector, while the pushing distance is capped at $20$ mm for each action, consistent with the $d_{push}$ used in our \textit{Caging in Time} experiments.

We introduced two types of disturbances to simulate real-world scenarios with imperfect perception and communication delays. Firstly, positional noise was added as Gaussian noise with a standard deviation of the noise level in both $x$ and $y$ directions, simulating the imperfect perception, such as that from an RGBD camera under dim lighting conditions. Secondly, to emulate potential packet loss during data transmission or sudden obstruction of the object from the camera, we introduced a network lag that for each execution step, there was a $50\%$ possibility that the controller would receive the object's position from 0.5 or 1 second ago.

\begin{figure}[t]
   \centering \includegraphics[width=0.45
   \textwidth]{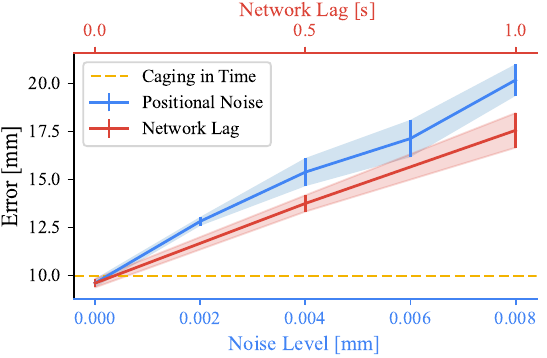}
   \caption{Evaluation results of the P-controller in terms of the noise level (blue) and network lag (red) compared to the \textit{Caging in Time} framework (yellow) under $20$mm cage size and 128 candidate actions.}
   \label{fig:quanti}
   \vspace{-5pt}
\end{figure}

The experimental results, as shown in Fig. \ref{fig:gauss} and Fig. \ref{fig:quanti}, demonstrate that our \textit{Caging in Time} approach achieves accuracy comparable to that of the P-controller with perfect perception. Moreover, the accuracy of the P-controller deteriorates significantly under the influence of perception noise or network lag. As shown in Fig. \ref{fig:gauss}, when the Gaussian noise is large, the resultant trajectory tends to exhibit random motion patterns, whereas network lag induces classic sinusoidal fluctuations in the trajectory. 

\begin{figure}[t]
   \centering \includegraphics[width=0.44\textwidth]{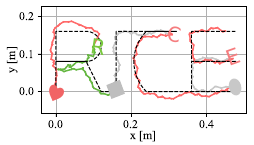}
   \caption{Trajectory following with in-task perturbation using six different objects (\textit{4, 7, 9, R, C, E}) for tracing ``RICE''. The manipulated objects are shown at the start point of each segment after the switch, with the color of the objects matched with their trajectories. The real recording is shown in Fig. \ref{fig:front}.}
   \label{fig:rss}
   \vspace{-10pt}
\end{figure}

These findings underscore the robustness of our \textit{Caging in Time} framework, particularly in real-world scenarios where perception and communication are often imperfect. The open-loop nature of our approach eliminates the need for continuous feedback, making it inherently resilient to perception noise and network delays that can significantly impair the performance of closed-loop methods.

\subsubsection{In-task Perturbations}

The robustness of \textit{Caging in Time} was further tested by maneuvering objects along specially designed trajectories shaped by the letters ``RICE'', as shown in Fig. \ref{fig:rss}. 
% This series of experiments was conducted using a variety of objects, including the nine shapes presented in Fig. \ref{fig:objs}, as well as character-like shapes such as ``R'' and ``S''. 
Notably, during this task, objects were intermittently randomly replaced by a human operator to test the system's ability to adapt to new shapes and position perturbations while maintaining trajectory precision in one single task. The results have further shown the robustness of our proposed \emph{Caging in Time} framework: as long as the new object was positioned within the current PSS $\mathcal{Q}_t$, the robot was able to consistently cage and manipulate it along the reference trajectory, demonstrating the method's robustness and adaptability.

\subsection{Dynamic Tasks}

To evaluate the capabilities and limitations of \textit{Caging in Time} in handling various dynamic tasks, we conducted a series of experiments using a tennis ball of diameter $6.6$cm with end-effector plates in different sizes, as shown in Fig. \ref{fig:setup}.

\begin{figure}[t]
   \centering \includegraphics[width=0.5
   \textwidth]{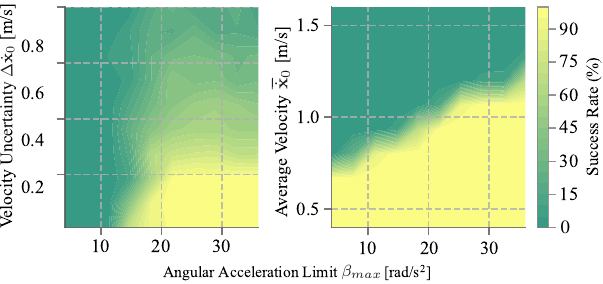}
   \caption{Ball catching success rate evaluation of finding feasible action sequences in PSS propagation over initial velocity uncertainty range $\Delta\dot{\mathbf{x}}_0$, average initial ball velocity $\overline{\dot{\mathbf{x}}}_0$, and hardware limitation represented by the end-effector angular acceleration limit $\beta_{max}$.}
   \label{fig:acclimit}
   \vspace{-10pt}
\end{figure}

\begin{figure*}[t]
\centering
\includegraphics[width=0.999\textwidth]{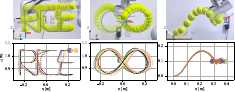}
\vspace{-20pt}
\caption{Experimental validation of \textit{Caging in Time} for dynamic manipulation tasks using a tennis ball and a support surface where dimension $n=1$ (See Fig. \ref{fig:setup}). \textit{Top row}: Representative frames from video recordings, with overlaid transparent balls showing the ball's position at different time points. \textit{Bottom row}: Trajectory plots of 10 repeated trials for each task. \textit{Left}: Ball balancing while tracing ``RICE''. \textit{Middle}: Ball balancing along an $\infty$-shaped trajectory. \textit{Right}: Ball catching with the balls rolling down from the same height on a slope to achieve consistent initial tossing velocity.}
\label{fig:dy2D}
\end{figure*}

Like quasi-static tasks, dynamic experiments also focused on three key aspects: 1) Can \textit{Caging in Time} achieve robust ball balancing on a moving plate along different trajectories without real-time sensing feedback? 2) How well does the framework perform in more extreme dynamic scenarios, such as catching a ball in an open-loop manner? 3) How robust is \textit{Caging in Time} when handling uncertainties higher-dimensional state spaces?

In the following experiments, the plate followed a predefined translational trajectory $\mathcal{T} = \{\mathbf{x}_p(t)\}_{t=0}^T$. While the trajectory of the plate was preset, the tilt angle of the plate at each time step is computed by our framework. A Cartesian space motion control is then used to control the robot's end-effector to follow this trajectory in a pure open-loop manner. For all dynamic experiments, we also used MAE to quantify the performance. Since the plate accurately tracks the target trajectory, the MAE is calculated as the average distance between the ball and the center of the plate across all time steps and all trials. The computation time for each step in action selection and PSS propagation was $67.3 \pm 12.6 $ ms.

\subsubsection{Dynamic Sensitivity Analysis}

Before proceeding with experiments, we conducted a dynamic sensitivity analysis to examine how \textit{Caging in Time} performance in the ball catching tasks responds to initial state distribution and hardware limitations, as previously mentioned in Alg. \ref{alg:caging_in_time} and Eq. \ref{eq:qp}. Fig. \ref{fig:acclimit} illustrates the success rates for finding feasible action sequences in Alg. \ref{alg:caging_in_time} for a ball catching task across varying initial velocity uncertainties $\Delta\dot{\mathbf{x}}_0$, average initial velocities $\overline{\dot{\mathbf{x}}}_0$, and end-effector angular acceleration limits $\beta_{max}$. The left plot fixes $\overline{\dot{\mathbf{x}}}_0$ at 0.8m/s, demonstrating how acceleration capabilities counteract uncertainty effects, while the right plot fixes $\Delta\dot{\mathbf{x}}_0$ at 0.05m/s, revealing velocity-hardware requirement relationships. We sampled each parameter using 10 evenly-spaced values within their ranges, with 100 trials per configuration done in simulated PSS propagation in Alg. \ref{alg:caging_in_time}.

As shown in Fig. \ref{fig:acclimit}, with a fixed $\overline{\dot{\mathbf{x}}}_0$ of 0.8 m/s, the tolerance of $\Delta\dot{\mathbf{x}}_0$ improved to 0.2 m/s as $\beta_{max}$ increased beyond 20 rad/$\text{s}^2$, indicating that higher acceleration capabilities allow for the system to accommodate larger PSS. Similarly, with $\Delta\dot{\mathbf{x}}_0$ fixed at 0.05 m/s, higher $\overline{\dot{\mathbf{x}}}_0$ required proportionally higher $\beta_{max}$ to maintain 100\% success rates. Based on our Franka robot's maximum end-effector angular acceleration, corresponding to the robot configuration used for this task, of approximately 25 rad/$\text{s}^2$, we determined our setup could reliably handle initial velocities up to 0.9 m/s with uncertainty ranges below 0.15 m/s, which guided our subsequent experimental design for dynamic tasks.

\subsubsection{Ball Balancing}

We first tested ball balancing with dimension $n=1$ on two distinct trajectories, as shown in Fig.~\ref{fig:dy2D}. The plate used in this task is 16cm long and has a guide rail to constrain the ball's rolling to the x-axis only, as shown on the bottom right of Fig. \ref{fig:setup}. The ``RICE'' trajectory, featuring sharp turns, resulted in an average error of $20.12 \pm 5.66$ mm. The $\infty$-shaped trajectory, repeated four times to test smooth recurring curves, yielded an average error of $26.59 \pm 7.54$ mm. Both experiments were repeated 20 times, with the ball consistently remaining on the plate, demonstrating the robustness and stability of \textit{Caging in Time} for long-horizon dynamic tasks. 

We observed that sharp turns in the plate's trajectory led to sudden changes in $\ddot{\mathbf{x}}_p$, increasing the control difficulty and causing larger errors in the ball's trajectory. Additionally, when tracking the $\infty$-shaped trajectory, $\ddot{\mathbf{x}}_p$ was non-zero for most of times, resulting in a persistent non-zero tracking error. This led to a higher average error for the $\infty$-shaped trajectory compared to the ``RICE'' trajectory, even though the maximum positional error of the ball's trajectory was relatively smaller.

\subsubsection{Ball Catching}

To further challenge the capability of \textit{Caging in Time} in handling complex dynamic tasks, we implemented an open-loop ball catching experiment using the same plate with a 1D track constraint where dimension $n=1$. In actual implementation, to ensure appropriate timing for initiating the action sequence, we utilized OpenCV to detect when the ball entered the camera frame. The open-loop action was triggered upon detection, with a manually tuned fixed time offset. Within the \textit{Caging in Time} framework, we considered the task to start when the ball made contact with the plate, ignoring bouncing effects. Additionally, we incorporated a hard-coded translational retreat motion to minimize ball rebounds on the plate.

We conducted two sets of experiments. In the first experiment, balls were released from a fixed height on an inclined slope, ensuring consistent initial position and approximately same velocity of the ball upon leaving the slope. 
% Under these controlled conditions and with known release parameters, we tested the system's ability to catch the ball using the track-constrained plate in a fully open-loop manner. 
Out of 20 trials, the system successfully caught the ball in all 20 attempts, as shown in Fig.~\ref{fig:dy2D} (right). This demonstrates the reliability of \textit{Caging in Time} under well-defined initial conditions.
\begin{figure}[t]
   \centering \includegraphics[width=0.49
   \textwidth]{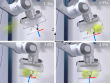}
   \vspace{-6pt}
   \caption{Demonstration of ball catching with human tossing using representative frames of 9 overlayed trials with time labels. Results along with Fig. \ref{fig:front} show that \textit{Caging in Time} is able to catch the ball tossed inaccurately by a human using the same open-loop action sequence in Fig. \ref{fig:dy2D}. }
   \label{fig:dyhum}
   \vspace{-10pt}
\end{figure}

In the second experiment, we introduced greater uncertainty by having a human manually toss the ball onto the plate, as illustrated in Fig.~\ref{fig:dyhum} and Fig.~\ref{fig:front}. Due to the inherent difficulty in controlling the landing position of hand-tossed balls, we only recorded trials where the ball successfully landed on the plate. Out of 20 such successful tosses, the system was able to catch and stabilize the ball in 13 trials. This partial success rate demonstrates the framework's robustness to velocity uncertainties within a certain range, as well as the limitations of the \textit{Caging in Time} framework's pre-configured error tolerance. This tolerance is constrained to ensure feasible actions within the robot's hardware torque limits, and some human tosses evidently introduced velocity variations beyond this tolerance range.

To quantitatively evaluate task completion rates under varying uncertainty conditions, we analyzed success rates in ball catching across different initial velocity distributions. Our data shows that slope dropping with velocities of $0.84 \pm 0.12$ m/s achieved a 100\% success rate, while human tossing with higher variability beyond the tolerance range ($0.93 \pm 0.28$ m/s) yielded a 65\% success rate. This real-world evaluation further validates our dynamic sensitivity analysis in Fig. \ref{fig:acclimit} and demonstrate the framework's performance boundaries under different uncertainty magnitudes.

\begin{figure}[t]
   \centering \includegraphics[width=0.48
   \textwidth]{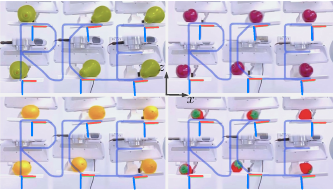}
   \caption{Demonstration of object-agnostic robustness in \textit{Caging in Time}. Balancing of diversely shaped YCB fruits from Fig. \ref{fig:setup} where dimension $n=1$ using the same open-loop action sequence for the tennis ball. Each figure shows multiple instances of the same fruit, representing its position at different time points during the balancing task.}
   \label{fig:dyobjs}
   \vspace{-12pt}
\end{figure}
\subsubsection{Caging with Uncertainty}

To test the framework's robustness to shape uncertainty, we balanced various fruits in YCB dataset shown in Fig. \ref{fig:setup} along the ``RICE'' trajectory with dimension $n=1$, as shown in Fig. \ref{fig:dyobjs}. Though the shape uncertainty brought higher uncertainties to the system dynamics, the framework could still make sure the object motion stays in the caged PSS. In 10 trials, all objects remained on the plate, demonstrating \textit{Caging in Time}'s adaptability to shape variations.

To note that, The success of these tasks can be partially attributed to the nature of irregular shapes, which are less likely to roll freely due to their inherent energy traps created by their non-uniform geometries. This characteristic actually aids in maintaining stability, as the objects tend to settle into local energy minima, complementing the caging strategy of our framework.

\subsubsection{Caging in Higher Dimensions}

\begin{figure}[t]
   \centering \includegraphics[width=0.33
   \textwidth]{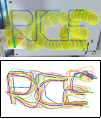}
   \vspace{-5pt}
   \caption{Ball balancing  using \textit{Caging in Time} with the dimension $n=2$ while tracing ``RICE''. \textit{Left}: Representative frames from recordings. The overlaid transparent balls show the ball's position at different time points during the task. \textit{Right}: Qualitative trajectory plots of 10 repeated trials for each task. }
   \label{fig:dy3D}
   \vspace{-10pt}
\end{figure}

Lastly, we extended our experiments to higher-dimensional spaces using a 16cm$\times$16cm plate, where the absence of the guide rail allows the ball to move freely in both X and Y directions, significantly increasing the complexity of the balancing task. As shown in Fig. \ref{fig:dy3D}, we performed ball balancing with the dimension $n=2$, where the ball's state is represented in a 4D state space $(x_t, y_t, \dot{x}_t, \dot{y}_t)$, two dimensions higher than all previous tasks. 

In 10 trials, the ball consistently remained on the plate, showcasing \textit{Caging in Time}'s applicability to higher-dimensional scenarios. The trajectories shown in Fig.~\ref{fig:dy3D} are for illustrative purposes due to non-orthogonal camera placement and may not represent exact quantitative performance, where we can visibly tell the trajectories exhibit larger deviations compared to the previous balancing experiments. This increased error is attributed to the unrestricted rolling direction in this setup, which introduces greater uncertainties and control challenges.

% \begin{figure}[h]
%    \centering \includegraphics[width=0.48\textwidth]{figs/RICERSS_.pdf}
%    \caption{Trajectory following with in-task perturbation. Six objects (4, 7, 9, R, C, E in Fig. \ref{fig:objs}) are used for ``RICE'' and four objects (2, 3, R, S) for pushing ``RSS''. On the left, the manipulated objects are shown at the start point of each segment after the switch, with the color of the objects matched with their trajectories. On the right, recording at different time for different objects are concatenated into a single figure.}
%    \label{fig:robust}
% \end{figure}

% Space for Results
% -----------------
% Include your results here once the experiments have been conducted.

\section{Applications in practice}
\label{sec:apps}

Applications for \textit{Caging in Time} could potentially extend beyond previous experiments to broader manipulation domains. This section explores the possibilities and practical considerations for implementing \textit{Caging in Time} in various scenarios.

\subsection{Example Applications}

In \textit{Caging in Time}, according to Sec. \ref{sec:general}, object states are represented through PSS $\mathcal{Q}_t$ within task-specific state spaces, with actions $u_t$ ensuring $\mathcal{Q}_{t+1} = \Pi(\mathcal{QV}_t) \subset \mathcal{S}^{t+1}_{cage}$, where $\mathcal{QV}_t$ is the motion bundle in Eq. \eqref{eq: QV_t} and $\Pi$ is the propagation function in Eq. \eqref{piact}. Below are tasks that we believe can be effectively represented and enhanced through this framework.

\subsubsection{In-Hand Manipulation} 
For in-hand manipulation, our framework enables instantaneously incomplete cages to dynamically form complete cages over time, guiding objects toward target states with containment guarantees, significantly relaxing the requirement for in-hand caging \citep{bircher_complex_2021} in both quasistatic and dynamic cases.

In quasi-static cases, the state space is $\mathcal{S}_{obj} \subset SE(3)$, encoding object pose. The cage can form in two ways: through geometric constraints, where spatial finger contact arrangements prevent the object from escaping, or through energy-based constraints using potential energy functions where the energy boundary forms the cage as defined in Eq.~\eqref{eq:energy}:
$\mathcal{S}_{cage}^t = \{\mathbf{q}_t \in SE(3) | E(\mathbf{q}_t) < E_{max}\},$
where $E(\mathbf{q}_t)$ represents the combination of gravitational potential energy and virtual potential fields that model contact constraints. These two approaches can also work in tandem -— potential fields control object motions while geometric constraints provide explicit caging boundaries, as demonstrated in \citep{bircher_complex_2021}

In dynamic cases, the state $\mathbf{q}_t$ becomes $(\mathbf{x}_t, \dot{\mathbf{x}}_t)$ and the state space expands to $\mathcal{S}_{obj} \subset SE(3) \times se(3)$ to incorporate both the pose and the twist, where the cage emerges as energy barriers. Similarly in Eq.~\eqref{eq:E_xt}, the energy-based cage can be explicitly defined as:
$\mathcal{S}_{cage}^t = \{(\mathbf{x}_t, \dot{\mathbf{x}}_t) \in SE(3) \times se(3) | E_{p}(\mathbf{x}_t) + E_{k}(\dot{\mathbf{x}}_t) < E_{max}\},$
where $E_{p}(\mathbf{x}_t)$ represents the potential energy from gravity and contact forces, and $E_{k}(\dot{\mathbf{x}}_t) = \frac{1}{2}\dot{\mathbf{x}}_t^T \mathbf{M} \dot{\mathbf{x}}_t$ is the kinetic energy determined by the object's velocity and inertia matrix $\mathbf{M}$. 

Maintaining this time-varying cage requires coordinated finger actions including sliding for continuous contact adjustment, rolling for smooth surface transitions, and gaiting for discrete contact reestablishment. These motion primitives and underlying finger reconfigurations strategically evolve both geometric constraints and energy barriers over time, known as \textit{Caging in Time}.

\subsubsection{Extrinsic Dexterity}
Extrinsic dexterity leverages environmental features as complementary cage elements, extending manipulation capabilities beyond what is possible with end-effectors alone.

As aforementioned, in quasi-static setup, the state space $\mathcal{S}_{obj} \subset SE(3)$ represents the object pose, while the cage is defined geometrically through the joint arrangement of the end-effector and environmental features like surfaces, corners, and obstacles. Similarly to Eq.~\eqref{eq:caging}, these elements collectively form spatial barriers around $\mathcal{C}_{free}^{obj}$, preventing uncontrolled object motion while considering the potential energy from gravity and interactions \citep{mahler2016energy}.

Also in dynamic scenarios, the state space expands to $\mathcal{S}_{obj} \subset SE(3) \times se(3)$. Similarly to the energy-based cage for in-hand manipulation, the cage can be defined as $\mathcal{S}_{cage}^t = \{(\mathbf{q}_t, \dot{\mathbf{q}}_t) \in SE(3) \times se(3) | E_{p}(\mathbf{q}_t) + E_{k}(\dot{\mathbf{q}}_t) < E_{max}\}$, where the kinetic energy of the controlled object momentum, the potential energy of the gravitational effects and the contact forces together create energy barriers that guide the motion of the object within admissible regions, allowing manipulation with fewer contacts.

The cage is maintained through strategic management of object-environment and object-robot contacts relative to environmental features, and multi-modal motion primitives like pushing, flipping and grasping. Unlike traditional extrinsic dexterity approaches \citep{yang2024learningextrinsicdexterityparameterized}, \textit{Caging in Time} explicitly plans transitions between contact states without requiring continuous feedback, extensive training, or specific object geometry limits, ensuring continuous caging in time despite uncertainties.

\subsubsection{Deformable Object Manipulation}
For deformable objects, each state $\mathbf{q}_t \subset \mathbb{R}^3$ directly represents the object's continuous shape in $\mathbb{R}^3$ within the state space $\mathcal{S}_{obj} \subset 2^{\mathbb{R}^3}$, rather than a discrete configuration, naturally aligning with PSS representation. These shapes can be encoded through parametric approaches with predefined models \citep{pokorny2013grasp}, or nonparametric methods using point distributions \citep{shi2023robocook}. In cases where objects are considered non-elastic, they maintain their deformed shape after contact is removed, allowing representation within $\mathbb{R}^3$ without modeling time-dependent recovery dynamics.

The cage forms through strategically distributed contacts that constrain the PSS while transforming the object shape. Using an estimated deformation model that can cover general PSS motion similarly to Eq. \eqref{eq:Vqt} in planar pushing, we can create spatial barriers that limit possible deformations from exceeding the cage, enabling shape control without requiring precise physical models of complex deformation or large amounts of data.

Unlike object pushing, the action space for deformable object manipulation encompasses material-specific motion primitives such as pushing, pinching, folding, and rolling. By sequencing these primitives strategically, \textit{Caging in Time} creates virtual cages whose deformation directly controls object transformation, maintaining reliable manipulation even under occlusion or complex deformation scenarios.

\subsection{Implementation Guidelines}
Adapting \textit{Caging in Time} to new manipulation applications requires systematic consideration of several key elements:

\textit{State Space Formulation}: Define $\mathcal{S}_{obj}$ to capture task-relevant properties as mentioned above, along with state propagation function discussed in Sec. \ref{sec:sub_caging_in_time}:
\begin{equation}
    \begin{aligned}
        \mathcal{QV}_t &= \{(\mathbf{q}_t, \mathbf{v}_t) \mid \mathbf{q}_t \in \mathcal{Q}_t, \mathbf{v}_t \in U(\mathbf{q}_t, u_t)\}, \\
        \mathcal{Q}_{t+1} & = \Pi(\mathcal{QV}_t)\\
    \end{aligned}
\end{equation}
Note that the propagation function $\Pi$ can be implemented either through analytical models as presented in this work or data-driven predictive frameworks such as neural networks and diffusion-based models.

\textit{Action Space Design}: Define the action space $\mathcal{U}$ satisfying $\forall t, \exists u_t \in \mathcal{U} : \Pi(\mathcal{QV}_t) \subset \mathcal{S}^t_{cage}$. The action space can be discrete as in our pushing tasks or continuous as in our ball balancing task with tilting angles, depending on task requirements and the control strategy. 

\textit{Cage Region}: Design the cage in time $\mathcal{S}^t_{cage}, t=0,1,\dots, T$ to balance precision and robustness while ensuring the caging condition $\mathcal{Q}_{t+1} = \Pi(\mathcal{QV}_t) \subset \mathcal{S}^t_{cage}$ holds. While our work mainly employed shape invariant cages, \textit{Caging in Time} naturally supports time-varying cage morphology where $\mathcal{S}^t_{cage}$ can reshape to accommodate state-dependent constraints, enabling wider applications such as in-hand manipulation or deformable object manipulation.

\textit{Action Selection}: Determining optimal actions $u_t \in \mathcal{U}$ requires a strategy that satisfies $\Pi(\mathcal{QV}_t) \subset \mathcal{S}^t_{cage}$. Our approach employed optimization-based methods like exhaustive search for planar pushing or quadratic programming for ball manipulation. Alternative paradigms such as reinforcement learning could extend applicability to more scenarios. Importantly, as discussed in Alg. \ref{alg:caging_in_time} and Fig. \ref{fig:acclimit}, hardware constraints must be incorporated to ensure the actions can be deployed in real-world physical systems.

\textit{Computational efficiency}: Current action selection and PSS propagation require $26.2 \pm 8.5$ ms (pushing) and $67.3 \pm 12.6$ ms (ball balancing) on a single CPU thread, with potential for optimization through GPU acceleration. Though currently offline-computed, these timings show feasibility for integrating \textit{Caging in Time} into the online planning for more diverse and dynamic scenarios.
\section{Conclusion} 
\label{sec:conclusion}

In this work, we proposed and evaluated \textit{Caging in Time}, a novel theory for robust object manipulation. Our framework demonstrated robust performance in both quasi-static and dynamic tasks without requiring detailed object information or real-time feedback. Rigorous evaluations highlighted the framework's resilience and adaptability to various objects and dynamic scenarios. The \textit{Caging in Time} approach proved effective in handling new objects, positional perturbations, and challenging dynamic tasks, showcasing its potential for reliable manipulation in uncertain environments.

While the current \textit{Caging in Time} framework shows promising results, it is important to acknowledge its limitations. Presently, the framework requires manual definition of task-specific parameters and the PSS propagation function. This reliance on human expertise may limit its generalizability to a wider range of manipulation tasks. Additionally, the current approach may not fully capture the complexity of certain real-world scenarios where object interactions and environmental factors are highly unpredictable.

Looking forward, we aim to address these limitations and expand the horizons of \textit{Caging in Time}. A key direction for future research is the integration of learning techniques, LLMs, and diffusion models to autonomously construct and learn \textit{Caging in Time} tasks. This approach could enable the framework to automatically derive appropriate propagation functions and strategies, reducing the need for manual parameter tuning. We also plan to explore its applications as mentioned in Sec. \ref{sec:apps}, while potentially leveraging reinforcement learning to handle increased task complexity and environmental variability.

\begin{acks}
This work was supported by the National Science Foundation under grant FRR-2240040.
\end{acks}

%%Harvard (name/date)
\bibliographystyle{SageH}
%%Vancouver (numbered)
%\bibliographystyle{SageV}
\bibliography{Caging_ref}

\section*{Appendix}
\label{sec:appendix}

\subsection{PSS Propagation for Quasi-static Pushing}
\label{sec:PeskinPush}
As shown in Fig.~\ref{fig:POA1}, assuming the line pusher is sufficiently long,
we denote $d_{con} \leq d_{push}$ as the moving distance of the line pusher after contacting the bounding circle of the object.
With the assumption that the motion of the object is quasi-static, we can apply the principles of energy conservation and force equilibrium to bound the displacement of the object.
The force equilibrium condition yields that the frictional force is equal to the pushing force, i.e., $F_{\mathrm{friction}} = F_{\mathrm{push}}$.
The work done by the push $W_{\mathrm{push}}$ is partially absorbed by the translational friction energy $E_{\mathrm{loss}}$, therefore, we have $W_{\mathrm{push}} = F_{\mathrm{push}}\cdot d_{con} \geq F_{\mathrm{friction}} \cdot \lVert \Delta q_t \rVert = E_{\mathrm{loss}}$. This gives a bound for the magnitude of the object's displacement $\lVert \Delta q_t \rVert \leq d_{con}$.

Additionally, through further analysis using the Peshkin distance~\citep{peshkin_motion_1988}, we can further bound the side displacement of the object $\lVert \Delta {q_t}_\parallel \rVert$, which is the displacement of the object parallel to the line pusher.

\subsubsection{The Peshkin Distance}

\begin{figure}[h]
   \centering \includegraphics[width=0.24\textwidth]{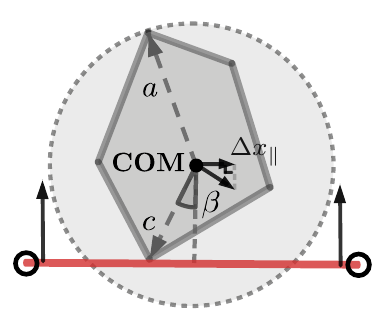}
   \caption{The illustration of notations used for Peshkin distance calculation. 
   The Peshkin distance bounds the side displacement of the object by $m / 2$, where $m$ denotes the pushing distance (same as $d_{con}$ in Sec.~\ref{sec:poa_prop}).}
   \label{fig:Peskin}
\end{figure}

The Peshkin distance defines the minimal translational displacement of a fence necessary to achieve alignment of an object's edge with the fence. This distance is a function of the object's rotation center, corresponding to the sticking condition that results in the minimum rotation speed. For a reorientation process, as shown in Fig.~\ref{fig:Peskin}, given the sticking-slowest rotation scenario, the Peshkin distance $m$ is determined by:
\begin{equation}
m=\frac{a^2+c^2}{2c}\left(\ln \left|\frac{1-\cos \beta_{1}}{1+\cos \beta_{1}}\right| - \ln \left|\frac{1-\cos \beta_{0}}{1+\cos \beta_{0}}\right|\right),
\end{equation}

where $a$ represents the radius of the object's bounding circle centered at the mass center, $c$ is the distance from the mass center to the contact point, and $\beta$ is the angle subtended by the motion direction and the line connecting the contact point with the mass center. The angles $\beta_{0}$ and $\beta_{1}$ correspond to the initial and final orientation angles for this rotation.

To reverse engineer this process and compute the maximum change in $\beta$ (denoted as $\Delta \beta$) for a fixed push distance $m$ (i.e., $d_{con}$ in Sec.~\ref{sec:poa_prop}), the following derivation is used:
\begin{equation}
e^{\frac{cm}{a^2+c^2}} = \frac{\tan \frac{\beta_1}{2}}{\tan \frac{\beta_0}{2}} = \frac{\tan \frac{\beta_0 + \Delta \beta}{2}}{\tan \frac{\beta_0}{2}} \approx 1 + \csc \beta_0 \Delta \beta
\end{equation}

Assuming that the push distance $m$ is much smaller than both $a$ and $c$, we approximate:
\begin{equation}
e^{\frac{cm}{a^2+c^2}} \approx 1 + \frac{cm}{a^2+c^2}
\end{equation}

Consequently, the maximum change in $\beta$ can be approximated by:
\begin{equation}
\Delta \beta = \frac{c \sin \beta_0}{a^2+c^2} m
\end{equation}

These derivations provide a mathematical basis for predicting the rotational effect of a planar pushing interaction on an object, under the assumption of no slip at the contact point.

We can further get the bound of the side-displacement $\Delta x_{\parallel}$ (corresponding to $\lVert \Delta {q_t}_\parallel \rVert$):
\begin{equation}
\begin{aligned}
\Delta x_{\parallel} &= c \Delta \beta_0 \cos{\beta_0}\\
&=\frac{mc^2}{a^2+c^2}\sin{\beta_0}\cos{\beta_0}\leq m\sin{\beta_0}\cos{\beta_0}\leq \frac{m}{2}
\end{aligned}
\end{equation}

Therefore, the side displacement $\lVert \Delta {q_t}_\parallel \rVert$ is bounded by $\lVert \Delta {q_t}_\parallel \rVert \leq d_{con} / 2$. This bound plus the other bound $\lVert \Delta q_t \rVert \leq d_{con}$ constrain all the possible displacements of the object $\mathcal{V}_{q_t}$ to be inside a semi-ellipse shown by the dark green region on the right side of Fig.~\ref{fig:POA1}, for which the mathematical expression is given in Eq.~\eqref{eq:Vqt1}.

In summary, combining both the cases where the bounding circle of the object collides with the pusher or not, the set of all possible motions for a specific configuration $q_t$ is given by the following equations, i.e., an explicit expression of the abstract function $U$ defined in Eq.~\eqref{eq:U}.
\begin{equation}
\label{eq:Vqt1}
    \begin{aligned}
        &\mathcal{V}_{q_t} = U(q_t, u_t) = \text{if $\mathbf{d}(q_t, \mathcal{L}) > r + d_{push}$ or $u_t$ is \textit{None}:}\\
        & \quad \left\{(0, 0)\right\}\\
        &\text{otherwise:}\\
        & \quad \left\{ v_t = R(u_t) \begin{pmatrix} \Delta x \\ \Delta y \end{pmatrix} \bigg| \frac{\left(\Delta x\right)^2}{d_{con}^2} + \frac{\left(\Delta y\right)^2}{\left(\frac{d_{con}}{2}\right)^2} \leq 1, \Delta x \leq 0 \right\}
        % &\left\{\begin{array}{l}
        % \left\{(0, 0)\right\} \quad \text{if $\mathbf{d}(q_t, \mathcal{L}) > r + d_{push}$}\\
        % \left\{ v_t = R(u_t) \begin{pmatrix} \Delta x \\ \Delta y \end{pmatrix} \mid  \frac{\left(\Delta x\right)^2}{d_{con}^2} + \frac{\left(\Delta y\right)^2}{\left(\frac{d_{con}}{2}\right)^2} \leq 1, \Delta x \leq 0\right\}
        % \end{array}\right.
    \end{aligned}
\end{equation}

In the equation, $R(u_t) \in SO(2)$ is a rotation matrix of $u_t$.

\subsection{Dynamic Control Implementation Details}
\label{app:dynamic_control}

Here we provide additional details on the implementation of the dynamic control strategy.  

\subsubsection{System Dynamics}
\label{sec:appendix_dynamics}

The system dynamics of the ball on the plate can be described by the following equation:
\begin{equation}
\begin{aligned}
\ddot{\mathbf{x}}_t&= f(\mathbf{q}_t, u_t, \eta_m, \boldsymbol{\eta}_p, \eta_\mu)\\
&= \frac{m(1 + \eta_m)}{m + \frac{I_b}{r_b^2}} \Big(
    (I_{n+1} - \mathbf{n}(\boldsymbol{\theta})\mathbf{n}(\boldsymbol{\theta})^T) \\
    &\quad \times (\mathbf{g}_\theta + \mathbf{a}_p+ \boldsymbol{\eta}_p)
    \Big) - (\mu_r + \eta_\mu) \dot{\mathbf{x}}_{t} 
\end{aligned}
\end{equation}
where $I_{n+1}$ is the identity matrix, $\mu_r$ is the estimated rolling friction coefficient, and $\mathbf{n}(\boldsymbol{\theta}) \in \mathbb{R}^{n+1}$ is the normal vector of the plate as a function of  $\boldsymbol{\theta}$. The parameters $m$, $I_b$, and $r_b$ represent the mass, moment of inertia, and radius of the ball respectively. $\mathbf{g}_\theta$ is the gravitational acceleration vector in the plate frame, and $\mathbf{a}_p$ is the acceleration of the plate.

The terms $\eta_m \in \mathbb{R}$, $\boldsymbol{\eta}_p \in \mathbb{R}^{n+1}$, and $\eta_\mu \in \mathbb{R}$ represent random distributions accounting for uncertainties in mass, plate acceleration, and friction coefficient respectively. These uncertainties follow Gaussian distributions:

\begin{equation}
\begin{aligned}
\eta_i &\sim \mathcal{N}(0, \sigma^2_i), \quad i \in \{m, \mu\} \\
\boldsymbol{\eta}_p &\sim \mathcal{N}(\mathbf{0}, \Sigma_p)
\end{aligned}
\end{equation}
where $\Sigma_p \in \mathbb{R}^{(n+1) \times (n+1)}$ is the covariance matrix.

To derive the probability distribution of the ball's acceleration, $p(\ddot{\mathbf{x}}_t)$, we consider the system dynamics equation as a function of the random variables $\eta_m$, $\boldsymbol{\eta}_p$, and $\eta_\mu$. Given that these uncertainties follow Gaussian distributions, $\ddot{\mathbf{x}}_t$ can be approximated as a Gaussian distribution as well. We can express this distribution as:
\begin{equation}
\ddot{\mathbf{x}}_t \sim \mathcal{N}(\boldsymbol{\mu}_{\ddot{\mathbf{x}}}, \boldsymbol{\Sigma}_{\ddot{\mathbf{x}}})
\end{equation}
where $\boldsymbol{\mu}_{\ddot{\mathbf{x}}}$ is the mean acceleration vector and $\boldsymbol{\Sigma}_{\ddot{\mathbf{x}}}$ is the covariance matrix. The mean $\boldsymbol{\mu}_{\ddot{\mathbf{x}}}$ is obtained by setting all uncertainties to their expected values (zero) in the system dynamics equation. 

For the special case where $n=1$, we can reformulate the system dynamics into a control-affine form to facilitate the application of Control Barrier Functions (CBFs) and Control Lyapunov Functions (CLFs):
\begin{equation}
\ddot{x}_t = f_t + g_t u
\end{equation}
where:
\begin{equation}
\begin{aligned}
f_t =& \frac{m}{m + \frac{I_b}{r_b^2}} (1 + \eta_{m})((g + (\ddot{z}_p+\eta_{z}))\sin\theta +\\
&(\ddot{x}_p+\eta_x)\cos\theta) - (\mu_r+\eta_{\mu}) \dot{x}_t \\
g_t =& \frac{m}{m + \frac{I_b}{r_b^2}} (1 + \eta_{m})((g + (\ddot{z}_p+\eta_{z}))\cos\theta - \\
&(\ddot{x}_p+\eta_x)\sin\theta)
\end{aligned}
\end{equation}

Here, $\eta = \{\eta_m, \eta_x, \eta_z, \eta_\mu\}$ represents random distributions accounting for uncertainties in the system. This representation separates the dynamics into a drift term $f_t$ (system behavior without control input) and a control term $g_t u$. 

This form simplifies the calculation of Lie derivatives for CBFs and CLFs. It allows for explicit representation of uncertainties in different system components through the $\eta$ terms, and decouples the control input $u=d\theta$, facilitating analysis of how control changes affect the system dynamics.

In practice, the variances of these uncertainties are manually determined. Larger variances indicate higher tolerance for errors, but also increase the difficulty of finding feasible actions.

\subsubsection{CBF and CLF}
\label{sec:cbfclf}
The CLF is defined as:
\begin{equation}
V(\mathbf{x}) = \sum_{\mathbf{q}_t} \mathcal{I}_t(\mathbf{q}_t)E(\mathbf{q}_t,\mathbf{g}_{\theta_t}, \mathbf{a}_p) - k_S S(\mathbf{x})
\end{equation}
where $\mathcal{I}_t(i,j)$ is the probability at the grid point $(i,j)$ in the discretized probability distribution, and $S(\mathbf{x})$ is the system entropy. The CLF is chosen as a weighted sum of the system's energy and negative entropy to balance the goals of minimizing energy expenditure and maintaining a well-distributed probability state, ensuring both stability and robustness in the face of uncertainties.

The entropy $S(\mathcal{Q}_t)$ is calculated as:
\begin{equation}
\label{eq:entropy}
S(\mathcal{Q}_t) = -\sum_{\mathbf{q}_t} \mathcal{I}_t(\mathbf{q}_t) \log \mathcal{I}_t(\mathbf{q}_t)
\end{equation}
The Lie derivatives for the CBF and CLF are calculated as:
\begin{equation}
\label{eq:lied}
\begin{aligned}
L_f h(\mathbf{x}) &= \frac{\partial h}{\partial \mathbf{x}} f(\mathbf{x}), L_g h(\mathbf{x}) = \frac{\partial h}{\partial \mathbf{x}} g(\mathbf{x}) \\
L_f V(\mathbf{x}) &= \frac{\partial V}{\partial \mathbf{x}} f(\mathbf{x}), L_g V(\mathbf{x}) = \frac{\partial V}{\partial \mathbf{x}} g(\mathbf{x})
\end{aligned}
\end{equation}
where $\frac{\partial h}{\partial \mathbf{x}}$ and $\frac{\partial V}{\partial \mathbf{x}}$ are the gradients of $h$ and $V$ with respect to the state $\mathbf{x}$, respectively.
In practice, these derivatives are approximated numerically due to the complexity of the expressions for $h$ and $V$.

\subsubsection{Numerical Implementation}

The numerical implementation of the control strategy is outlined in Alg. \ref{alg:dynamic_control}.

\begin{algorithm}[t]

\SetKwInput{KwData}{Input}
\SetKwInput{KwResult}{Output}
\caption{DynamicControl($\cdot$)}
\label{alg:dynamic_control}
\KwData{Initial state $\mathcal{Q}_0$, desired trajectory $\mathcal{T}=\{\mathbf{x}_p(t)\}_{t=0}^T$, time step $\Delta t$}
\KwResult{Success or failure of the task}
\For{$t = 0, \ldots, T-1$}{
$S(\mathcal{Q}_t) \gets$ EstimateEntropy($\mathcal{Q}_t$)\hfill \Comment{Eq.~\eqref{eq:entropy}}\\
Compute $E_{max}(\mathbf{g}_{\theta_t}, \mathbf{a}_p)$ and $E(\mathbf{q}_t,\mathbf{g}_{\theta_t}, \mathbf{a}_p)$ for all states in $\mathcal{Q}_t$\hfill \Comment{Eq.~\eqref{eq:energy}~\eqref{eq:E_xt}}\\
Calculate $L_f h$, $L_g h$, $L_f V$, $L_g V$\hfill \Comment{Eq.~\eqref{eq:lied}}\\
$d\boldsymbol{\theta} \gets$ SolveQP($L_f h, L_g h, L_f V, L_g V$) \hfill \Comment{Eq.~\eqref{eq:qp}}\\
$\boldsymbol{\theta}_{t+1} \gets \boldsymbol{\theta}_t + d\boldsymbol{\theta}$\\
$\mathcal{Q}_{t+1} \gets f(\mathcal{Q}_t, \boldsymbol{\theta}_{t+1})$ \hfill \Comment{ Eq.~\eqref{eq:probqp}}\\
\If{$\max_{\mathbf{q}_t} E(\mathbf{q}_t,\mathbf{g}_{\theta_{t+1}}, \mathbf{a}_p) > E_{max}(\mathbf{g}_{\theta_{t+1}}, \mathbf{a}_p)$}{
\Return{false}
}
}
\Return{true}
\end{algorithm}

This algorithm implements the control strategy by iteratively solving the QP and updating the system state. It ensures that the ball remains within the energy-based cage while following the desired trajectory. 

% The use of OSQP solver allows for efficient QP solutions at each time step, making real-time control feasible.
\end{document}